\documentclass[11pt]{article}
\usepackage{acl}
\usepackage{times}
\usepackage{latexsym}
\usepackage[T1]{fontenc}
\usepackage[utf8]{inputenc}
\usepackage{microtype}
\usepackage{inconsolata}
\usepackage{graphicx}
\usepackage{booktabs}
\usepackage{multirow}
\usepackage{amsmath}
\usepackage{array}
\usepackage{arydshln}
\usepackage{enumitem}
\usepackage{amssymb}
\usepackage{pifont}
\usepackage[breakable]{tcolorbox}
\usepackage{xcolor}
\usepackage{soul}
\definecolor{OnePassColor}{HTML}{6495ED}
\definecolor{TwoPassColor}{HTML}{F4A261}
\definecolor{DBias}{HTML}{2E7D32}
\definecolor{RBias}{HTML}{C62828}
\definecolor{SevMislabel}{HTML}{E65100}
\definecolor{InjectFail}{HTML}{6A1B9A}
\definecolor{AspectMis}{HTML}{1565C0}

\title{Beyond Scalar Scores: Exploring LLM-based Metrics for Clinical Significance Evaluation in Radiology Reports}

\author{Qingyu Lu$^{1}$\thanks{~~Equal contribution.} \quad Ruochen Li$^{2}$\footnotemark[1] \quad Liang Ding$^{3}$ \quad Yufei Xia$^{4}$ \quad Youxiang Zhu$^{5}$ \quad Dacheng Tao$^{1}$ \\
  $^{1}$Nanyang Technological University \quad $^{2}$Technical University of Munich \quad $^{3}$Alibaba \\
  $^{4}$University of Glasgow \quad $^{5}$University of Massachusetts Boston \\
  \texttt{qingyu.lu.ai@gmail.com}}

\begin{document}
\maketitle

\begin{abstract}
Reliable evaluation of generated radiology reports requires strict clinical accuracy, as omitted critical findings or mischaracterized radiographic observations can directly affect patient care. 
Existing metrics obscure this requirement by reducing report quality to a medically ungrounded scalar.
Although Large Language Models (LLMs) possess rich medical knowledge, they likewise struggle to draw a reliable boundary between clinically significant errors and harmless variation.
We study this boundary using \textbf{ReEvalMed} benchmark as testbed and evaluate metric-level clinical significance from detecting true clinical errors ("\emph{Discrimination}") and tolerating insignificant variations ("\emph{Robustness}").
Across 8 LLM evaluators under one-pass and two-pass settings, we identify a widespread \emph{discrimination bias}: models effectively detect errors but also over-penalize harmless rephrasings.
To mitigate this, we synthesize 4k report pairs and train lightweight interpretable metrics on Qwen3-8B and MedGemma-4B.
Our trained metric sharpens the clinical significance boundary, surpassing 32B-scale medical LLMs and remaining competitive with proprietary models. 
Crucially, the more costly two-pass setting fails to consistently improve overall performance and mainly trades discrimination for robustness. 
These findings suggest one-pass trained metrics as the practical choice for cost-sensitive deployment, with two-pass inference reserved for settings where D--R balance is critical. 
We will release the dataset and metric.
\end{abstract}

\section{Introduction}

\begin{figure}[t]
\centering
\includegraphics[width=0.85\columnwidth]{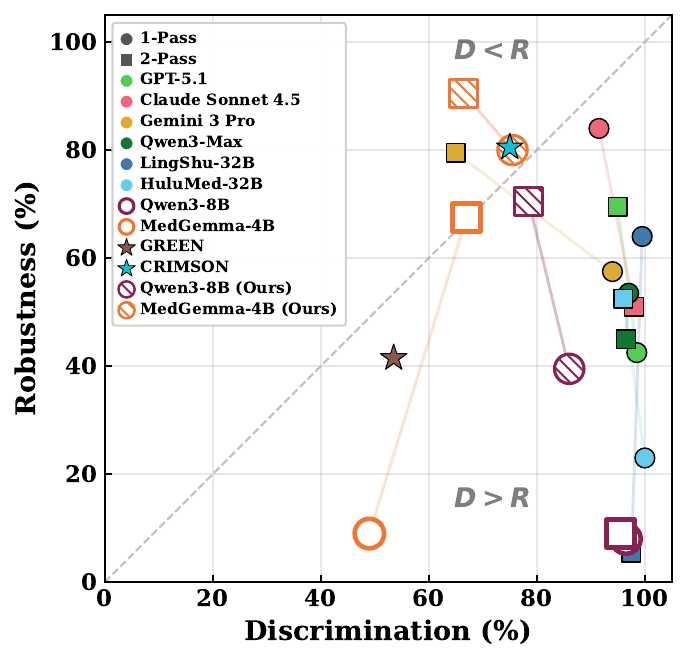}
\caption{Discrimination--Robustness accuracy of 8 LLM-as-evaluators, 2 medical metrics, and our post-trained models on ReEvalMed. \textit{Discrimination bias}: most LLMs fall below the diagonal (``D $>$ R'').}
\label{fig:dr_shift}
\end{figure}

Radiology reports describe imaging findings(e.g., lesion characteristics, anatomical abnormalities) that directly guide clinical diagnosis and treatment~\cite{tanno2025collaboration}, making clinically faithful evaluation essential for generated reports. 
Although vision-language models (VLMs) can generate such reports automatically from medical images~\cite{bannur2024maira,chen2024chexagent}, reliable evaluation remains an open challenge. 
A clinician-trusted evaluation metric should detect errors that could significantlyaffect clinical decisions while tolerating stylistic or clinically insignificant variation.
However, as recent study \cite{li2025reevalmed} demonstrate, both traditional lexical metrics and LLM-based medical report metrics that correlate strongly with human judgement (e.g. GREEN~\cite{ostmeier2024green} and RaTEScore~\cite{zhao2024ratescore}) often fail to separate clinically significant errors from minor variations. This boundary blurring undermines their reliability as indicators of clinical acceptability.

LLMs offer a path toward the interpretable evaluation that prior metrics lack: beyond scalar scoring, they can identify error spans, classify clinical error aspects, and assess clinical significance, thereby providing fine-grained feedback. 
However, empirical validation of LLMs as radiology report evaluators remains limited, and it is unclear whether their interpretive capacity translates into reliable clinical judgment. 
To this end, we study this question from two perspectives: the zero-shot behavior of current open-source and proprietary LLMs, and whether targeted data augmentation and fine-tuning can close the performance gap.

\begin{figure}[t!]
  \centering
  \includegraphics[width=0.9\columnwidth]{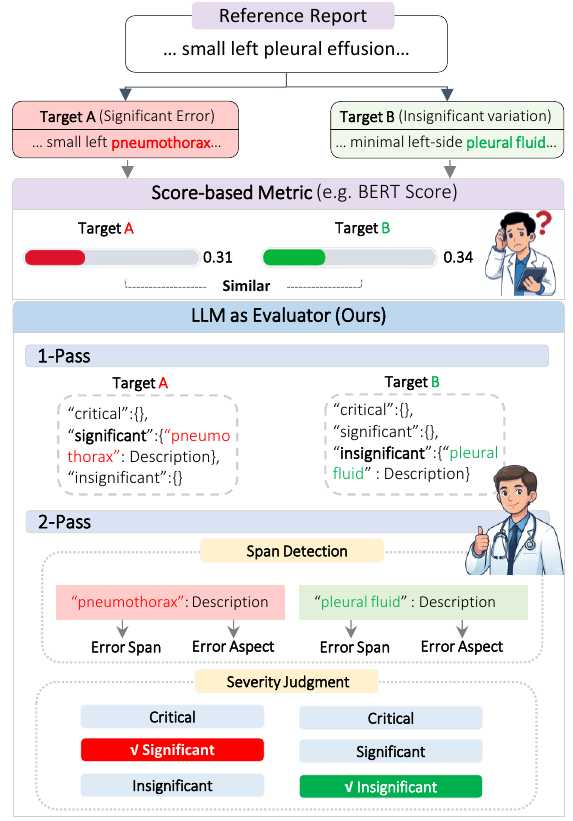}
  \caption{Score-based metrics (e.g.\ BERTScore) assign comparable scores to clinically significant errors and harmless rephrasings. Our LLM-as-evaluator outputs structured severity judgments, better distinguishing significant errors from insignificant variations.}
  \label{fig:overview}
\end{figure}

Motivated by recent LLM-based evaluators for generated texts \citep{lu2024error,kocmi2023gemba}, we design
one-pass and two-pass prompts for radiology evaluation (Figure~\ref{fig:overview}), where the former
directly outputs structured error annotations and the latter separates
\ding{192}~error span detection from \ding{193}~clinical significance judgment.
As shown in Figure~\ref{fig:dr_shift}, we observe a consistent \textbf{\textit{discrimination bias}}: \textit{current LLM evaluators struggle to distinguish clinically significant discrepancies from harmless report variations}, resulting in high discrimination
but low robustness accuracy.

To mitigate such discrimination bias and better delineate the boundary of clinical significance, we synthesize a balanced dataset reports annotated along 12 aspects and three error types (omission, fabrication, inaccuracy error) with Claude Sonnet, with clinician verification to ensure data quality.
To further validate the effectiveness of our synthesized dataset, we train a radiology report evaluation metric based on Qwen3-8B and MedGemma-4B. Our trained metric significantly improve the clinical significance boundary, outperforming 32B-scale medical LLMs and remaining competitive with proprietary LLMs. We further find that two-pass inference does not eliminate errors, but instead redistributes them between discrimination and robustness. Our contributions are three-fold:

\begin{itemize}[leftmargin=*,topsep=2pt,itemsep=1pt]
  \item \textbf{Analysis of LLM-based Evaluators.}
  We systematically evaluate 11 score-based metrics and 8 LLM-based metrics on ReEvalMed using one-pass and
  two-pass LLM prompting strategies, analyze the error pattern and revealing a consistent discrimination
  bias among most LLM metrics.
  \item \textbf{Clinically grounded data synthesis.}
  We synthesize a balanced dataset of 4k radiology report pairs covering all 12 aspects of ReEvalMed's error taxonomy (spanning omission, fabrication, and factual error), annotated at the span level and verified for clinical validity.
  \item \textbf{Lightweight interpretable metric.}
  Building on the synthesized data, we train Qwen3-8B via supervised fine-tuning (SFT) and reinforcement learning (RL) techniques using both one-pass and two-pass prompting format,
  achieving 78.5\% discrimination and 70.5\% robustness accuracy, surpassing medical LLMs at the 32B scale (such as Lingshu-32B~\cite{xu2025lingshu} and Hulu-Med-32B~\cite{jiang2025hulu}).
\end{itemize}

\begin{table*}[t]
\centering
\small
\renewcommand{\arraystretch}{1.6}
\setlength{\tabcolsep}{4pt}
\begin{tabular}{>{\centering\arraybackslash}m{3.0cm}>{\centering\arraybackslash}m{4.5cm}>{\centering\arraybackslash}m{5.5cm}>{\centering\arraybackslash}m{1.2cm}}
\toprule
\textbf{Criteria} & \textbf{Error Type} & \textbf{Error Aspect} & \textbf{\#} \\
\midrule
\multirow{2}{3.0cm}[-2ex]{\centering\textbf{Discrimination}} & \footnotesize Omission, Fabrication, Inaccuracy & \footnotesize Location, Severity, Description, Comparison/Progression & 120 \\
\cmidrule(l){2-4}
 & \footnotesize Inaccuracy Only & \footnotesize Negation, Modality, Size/Distance, Contradiction, Uncertainty, Terminology, Noise, Stylistic Variation & 80 \\
\midrule
\multirow{2}{3.0cm}[-2ex]{\centering\textbf{Robustness}} & \footnotesize Omission, Fabrication, Inaccuracy & \footnotesize Location, Severity, Description, Comparison/Progression & 120 \\
\cmidrule(l){2-4}
 & \footnotesize Inaccuracy Only & \footnotesize Negation, Modality, Size/Distance, Contradiction, Uncertainty, Terminology, Noise, Stylistic Variation & 80 \\
\bottomrule
\end{tabular}
\renewcommand{\arraystretch}{1}
\caption{ReEvalMed error taxonomy and test set composition. The benchmark spans 12 error aspects across two evaluation dimensions, \emph{Discrimination} (detecting clinically significant errors) and \emph{Robustness} (tolerating clinically insignificant variations), each with 200 samples (400 total).}
\label{tab:taxonomy}
\end{table*}

\section{Preliminary}
\label{sec:prelim}

\paragraph{Task formulation}
A radiology report evaluation metric takes a reference report ("\textbf{REF}") and a generated candidate report ("\textbf{TGT}") as input, and outputs a quality judgment.
Traditional metrics, including lexical metrics (BLEU~\cite{papineni2002bleu},
ROUGE-L~\cite{lin2004rouge}), embedding-based metrics
(BERTScore~\cite{zhang2019bertscore}), and clinical NLP metrics
(RadGraph~\cite{jain2021radgraph},
CheXbert~\cite{smit2020combining}), all reduce this judgment to a single \textbf{continuous score}.
However, a scalar score cannot distinguish clinically significant errors
from insignificant ones, as both may incur the same penalty under any
scalar metric.
We therefore adopt a \textbf{structured textual output} that explicitly labels each discrepancy as \emph{significant} or \emph{insignificant}, while also identifying its error span and clinical aspect. This provides fine-grained attribution beyond what scalar scoring can express.

\paragraph{ReEvalMed benchmark}
To assess whether a metric aligns with clinical judgment, 
ReEvalMed~\cite{li2025reevalmed} provides a fine-grained meta-evaluation benchmark for radiology report evaluation. 
Given paired reference and candidate reports with clinician-defined significance labels, ReEvalMed tests whether a metric can distinguish clinically significant errors from clinically insignificant variations. 
Its criteria were co-developed with clinicians to reflect clinically relevant discrepancies in radiology reports. 
ReEvalMed further organizes discrepancies along two axes: \emph{Error Type} (Omission, Fabrication, Inaccuracy) and \emph{Error Aspect} (e.g., Location, Severity, Negation), as summarized in Table~\ref{tab:taxonomy}. 
The test set contains 400 report pairs drawn from MIMIC-CXR~\cite{johnson2019mimic}.

\paragraph{Discrimination and Robustness}
ReEvalMed introduce two metrics along two complementary dimensions to measure the quality of evaluation:
\begin{itemize}[leftmargin=*,topsep=2pt,itemsep=1pt]
  \item \textbf{Discrimination} (200 test samples): the ability to detect errors that could materially affect diagnosis or treatment, such as changing lesion characterization from benign to malignant or substantially altering lesion size.

  \item \textbf{Robustness} (200 test samples): the ability to remain
  unpenalised by clinically harmless variations, such as synonymous wording, equivalent anatomical descriptions, or negligible measurement differences.
\end{itemize}
A clinically significance-sensitive metric should achieve high accuracy on both dimensions
simultaneously, i.e., correctly flagging clinically significant errors
while tolerating harmless variations.

\section{Analysis on LLM-based Metrics}
\label{sec:benchmark}

\subsection{LLM-as-Evaluator Methodology}

Motivated by recent LLM-based evaluators for generated
texts~\citep{lu2024error,kocmi2023gemba}, we design
structured prompts that instruct an LLM to compare a REF and
TGT pair and output fine-grained error annotations rather than a
scalar score.
Specifically, the evaluator identifies error spans and assigns each a
severity level from three classes:
\texttt{Critical} (internal contradictions that severely undermine report
trust), \texttt{Significant} (errors that meaningfully alter clinical
decision-making), and \texttt{Insignificant} (stylistic variations or
clinically harmless deviations).
We explore two inference paradigms for this task.

\paragraph{One-pass}
The LLM receives REF and TGT in a single prompt and directly outputs a
JSON object containing three severity
buckets (\texttt{critical}, \texttt{significant}, and
\texttt{insignificant}), each mapping error spans to their error aspects,
together with a free-text explanation.

\paragraph{Two-pass}
One-pass inference couples span detection and severity judgment into a
single output, which may cause compounding errors.
To address this, we decouple inference into two passes:
\begin{itemize}[leftmargin=*,topsep=2pt,itemsep=1pt]
  \item \textbf{Pass~1 (Span Detection)}: the LLM identifies all
  discrepancies between REF and TGT, outputting a JSON array of error
  spans with their aspects (e.g., \texttt{pneumothorax -- Description}).
  \item \textbf{Pass~2 (Severity Judgment)}: for each detected span, a
  second call outputs exactly one word (\texttt{Critical},
  \texttt{Significant}, or \texttt{Insignificant}), conditioned on
  aspect-specific criteria.
\end{itemize}

\paragraph{D/R classification rule}
Note that the three severity levels are \emph{error-level} labels
assigned to individual spans, whereas D and R are \emph{metric-level}
scores that measure how well a metric's predictions align with clinical
ground truth.
To bridge the two levels, we aggregate span-level severities into a
binary report-level prediction.
Denoting a TGT--REF pair as $(t, r)$, let $n_{\text{c}}$,
$n_{\text{s}}$, $n_{\text{i}}$ be the number of
\texttt{Critical}, \texttt{Significant}, and \texttt{Insignificant}
spans identified:
\begin{equation}
\setlength{\arraycolsep}{2pt}
\textsc{cls}(t,r)=
\begin{cases}
\textit{sig.} & n_c+n_s>0,\\
\textit{ins.} & n_c=n_s=0,\; n_i>0.
\end{cases}
\label{eq:cls}
\end{equation}
The \textbf{Discrimination score}~(D) is the accuracy of predicting
\textit{sig.}\ on the Discrimination subset, where all pairs
contain clinically significant errors; the \textbf{Robustness score}~(R)
is the accuracy of predicting \textit{insig.}\ on the Robustness
subset, where all pairs contain only clinically insignificant variations.

\subsection{Experimental Setup}

\begin{table}[t]
\centering
\footnotesize
\renewcommand{\arraystretch}{1.3}
\setlength{\tabcolsep}{3pt}
\begin{tabular}{cccccc}
\toprule
\multirow{2}{*}{\textbf{Type}} & \multirow{2}{*}{\textbf{Metric}} & \multicolumn{4}{c}{\textbf{Maximin Threshold}} \\
\cmidrule(lr){3-6}
 &  & \textbf{D\,($\uparrow$)} & \textbf{R\,($\uparrow$)} & \textbf{Avg\,($\uparrow$)} & \textbf{Gap\,($\downarrow$)} \\
\midrule
\multirow{3}{*}{NLP}
 & BLEU       & 19.0 & 19.0 & 19.0{\scriptsize\,($\pm$4.0)} & 0.0 \\
 & BERTScore  & 24.0 & 24.0 & 24.0{\scriptsize\,($\pm$4.2)} & 0.0 \\
 & AlignScore & 51.0 & 51.0 & 51.0{\scriptsize\,($\pm$5.0)} & 0.0 \\
\midrule
\multirow{4}{*}{Med.}
 & RadGraph   & 28.5 & 28.5 & 28.5{\scriptsize\,($\pm$4.5)} & 0.0 \\
 & RadBERTScore & 26.5 & 26.5 & 26.5{\scriptsize\,($\pm$4.2)} & 0.0 \\
 & RaTEScore  & 35.5 & 35.5 & 35.5{\scriptsize\,($\pm$4.8)} & 0.0 \\
 & CheXbert   & 47.0 & 49.0 & 48.0{\scriptsize\,($\pm$5.0)} & 2.0 \\
\midrule
\multirow{4}{*}{LLM}
 & GREEN      & 53.5 & 41.5 & 47.5{\scriptsize\,($\pm$4.8)} & 12.0 \\
 & RadFact    & 80.0 & 56.0 & 68.0{\scriptsize\,($\pm$4.5)} & 24.0 \\
 & FineRadScore & \textbf{86.5} & 68.5 & 77.5{\scriptsize\,($\pm$4.0)} & 18.0 \\
 & CRIMSON    & 75.0 & \textbf{80.5} & \textbf{77.8}{\scriptsize\,($\pm$4.2)} & 5.5 \\
\bottomrule
\end{tabular}
\caption{Comparison of score-based metrics on ReEvalMed. D and R scores are in accuracy (\%). Maximin threshold is applied to obtain balanced results. Best results are in \textbf{bold}. Numbers in parentheses denote 95\% bootstrap CIs for Avg.}
\label{tab:baseline}
\renewcommand{\arraystretch}{1.1}
\end{table}

\paragraph{Baselines}
We evaluate 11 score-based metrics that output continuous scores:
3 NLP metrics (BLEU~\cite{papineni2002bleu},
BERTScore~\cite{zhang2019bertscore}, and
AlignScore~\cite{zha2023alignscore});
4 medical metrics (RadGraph~\cite{jain2021radgraph},
RaTEScore~\cite{zhao2024ratescore}, CheXbert~\cite{smit2020combining}), and RadBERTScore, which replaces the generic BERTScore encoder with a radiology-domain encoder from RadEval~\cite{xu2025radeval};
and 4 LLM-based metrics (GREEN~\cite{ostmeier2024green}, which
prompts GPT-4 for finding-level error annotations, and
CRIMSON~\cite{baharoon2026crimson}, which fine-tunes MedGemma on
140K report pairs with GPT-5-generated severity labels via LoRA. FineRadScore~\cite{huang2024fineradscore} is an LLM-based line-by-line correction metric; RadFact~\cite{bannur2024maira} is an LLM-based entailment metric suite for radiology report evaluation, and we use its logical precision/recall scores. We instantiate RadFact and FineRadScore with GPT-5.1 for our experiments.

\paragraph{LLM-as-evaluators}
We evaluate 3 proprietary LLMs (GPT-5.1,
Claude Sonnet 4.5, Gemini 3 Pro) and
5 open-source LLMs (Qwen3-Max~\cite{yang2025qwen3},
LingShu-32B~\cite{xu2025lingshu},
Hulu-Med-32B~\cite{jiang2025hulu},
Qwen3-8B~\cite{yang2025qwen3}, and MedGemma-4B~\cite{sellergren2025medgemma}) using our one-pass and two-pass
prompts.

\paragraph{Prompt}
Prompt templates for both
\hyperlink{box:onepass}{one-pass} and
\hyperlink{box:twopass_p1}{two-pass} paradigms are provided in
Appendix~\ref{app:prompt}.

\paragraph{Hyperparameters}
All LLM model inferences use greedy decoding (temperature\,=\,0.0) with a maximum of 1024 output tokens.

\paragraph{Device and platform}
Proprietary models are accessed via their official APIs.
Open-source models ($\leq$32B) are served with
vLLM~\cite{kwon2023efficient} on a single NVIDIA A800 GPU.
The same device is used for all training experiments in
Section~\ref{sec:training}.

\begin{table*}[t]
\centering
\footnotesize
\renewcommand{\arraystretch}{1.2}
\setlength{\tabcolsep}{5pt}
\begin{tabular}{cccccccccc}
\toprule
\multirow{2}{*}{\textbf{Type}} & \multirow{2}{*}{\textbf{Model}} & \multicolumn{4}{c}{\textbf{1-Pass}} & \multicolumn{4}{c}{\textbf{2-Pass}} \\
\cmidrule(lr){3-6}\cmidrule(lr){7-10}
 &  & \textbf{D\,($\uparrow$)} & \textbf{R\,($\uparrow$)} & \textbf{Avg\,($\uparrow$)} & \textbf{Gap\,($\downarrow$)} & \textbf{D\,($\uparrow$)} & \textbf{R\,($\uparrow$)} & \textbf{Avg\,($\uparrow$)} & \textbf{Gap\,($\downarrow$)} \\
\midrule
\multirow{3}{*}{Proprietary}
 & GPT-5.1           & 98.5 & 42.5 & 70.5{\scriptsize\,($\pm$3.5)} & 56.0 & 95.0 & 69.5 & \textbf{82.3}{\scriptsize\,($\pm$3.5)} & 25.5 \\
 & Claude Sonnet 4.5 & 91.5 & \textbf{84.0} & \textbf{87.8}{\scriptsize\,($\pm$3.2)} & \textbf{7.5} & \textbf{98.0} & 51.0 & 74.5{\scriptsize\,($\pm$3.5)} & 47.0 \\
 & Gemini 3 Pro      & 94.0 & 57.5 & 75.8{\scriptsize\,($\pm$3.8)} & 36.5 & 65.0 & \textbf{79.5} & 72.3{\scriptsize\,($\pm$4.2)} & 14.5 \\
\midrule
\multirow{5}{*}{Open-source}
 & Qwen3-Max         & 97.0 & 53.5 & 75.3{\scriptsize\,($\pm$3.8)} & 43.5 & 96.5 & 45.0 & 70.8{\scriptsize\,($\pm$3.8)} & 51.5 \\
 & LingShu-32B       & 99.5 & 64.0 & 81.8{\scriptsize\,($\pm$3.3)} & 35.5 & 97.5 & 5.5 & 51.5{\scriptsize\,($\pm$2.0)} & 92.0 \\
 & Hulu-Med-32B      & \textbf{100.0} & 23.0 & 61.5{\scriptsize\,($\pm$3.0)} & 77.0 & 96.0 & 52.5 & 74.3{\scriptsize\,($\pm$3.8)} & 43.5 \\
 & Qwen3-8B          & 96.5 & 8.0 & 52.3{\scriptsize\,($\pm$2.3)} & 88.5 & 95.5 & 9.0 & 52.3{\scriptsize\,($\pm$2.5)} & 86.5 \\
 & MedGemma-4B       & 49.0 & 9.0 & 29.0{\scriptsize\,($\pm$4.0)} & 40.0 & 67.0 & 67.5 & 67.3{\scriptsize\,($\pm$4.8)} & \textbf{0.5} \\
\bottomrule
\end{tabular}
\caption{Comparison of LLM-as-evaluator on ReEvalMed benchmark. D and R scores are in accuracy (\%). Numbers in parentheses denote the half-width of the 95\% bootstrap CI for Avg. Best results are in \textbf{bold}.}
\label{tab:main}
\renewcommand{\arraystretch}{1}
\end{table*}

\subsection{Meta-Evaluation}

\paragraph{Score-based metrics}
To align the accuracy scale with LLM-based metrics, score-based metrics require a decision threshold to convert continuous values into binary D/R predictions.
We sweep all thresholds and apply the \textbf{maximin criterion}
$\max_\theta \min(D_\theta, R_\theta)$,
selecting the operating point that maximises the worse of the two dimensions\footnote{For FineRadScore~\cite{huang2024fineradscore} with severity-level labels, the selected threshold maps severity levels $<2$ to insignificant errors and levels $\geq2$ to significant errors.}.

\paragraph{Average and Gap}
We report \textbf{Avg}~$= (\text{D}+\text{R})/2$ for overall accuracy and
\textbf{Gap}~$= |\text{D}-\text{R}|$ for D--R imbalance.
An ideal evaluator achieves high Avg with low Gap. We report the 95\% confidence interval (CI) of Avg using bootstrap resampling over test cases. Detailed results in Appendix~\ref{app:stat_analysis}.

\subsection{Results}

\paragraph{Score-based metrics fail to distinguish significance}
Score-based metrics require a decision threshold to convert continuous
scores into binary significant/insignificant predictions.
We select the maximin threshold
$\max_\theta \min(D_\theta, R_\theta)$ that maximises the worse of the
two dimensions, balancing D and R accuracy
(see Appendix~\ref{app:curves}, Figure~\ref{fig:dr_curve_appendix}
for full trade-off curves).
As shown in Table~\ref{tab:baseline} (left), most non-LLM metrics remain
below 55\% on both D and R (e.g.\ BLEU 19.0/19.0, CheXbert 47.0/49.0),
indicating they cannot reliably separate significant errors from
harmless variations.
LLM-based score metrics perform notably better: CRIMSON achieves the
best maximin point (75.0/80.5), substantially outperforming GREEN
(53.5/41.5) and all non-LLM baselines.

\paragraph{LLM-as-Evaluators detect errors but over-penalize harmless variation}
Compared with score-based metrics, LLM-as-Evaluators achieve substantially higher Discrimination. As shown in Table~\ref{tab:main}, most LLMs reach D\,$\geq$\,94\% under 1-Pass, far exceeding the best score-based result (CRIMSON, 75.0\%). However, this sensitivity does not imply a reliable clinical-significance boundary. Robustness varies widely, ranging from 8.0\% (Qwen3-8B) to 84.0\% (Claude Sonnet 4.5), with most models below 60\%. Claude Sonnet 4.5 is the only model that performs strongly on both dimensions (D=91.5\%, R=84.0\%); all other LLM evaluators remain substantially imbalanced, typically detecting clinically significant errors while over-penalizing clinically insignificant variation.

\paragraph{1-Pass vs.\ 2-Pass}
Switching from 1-Pass to 2-Pass does not uniformly improve the Discrimination--Robustness balance. Instead, it shifts the bias pattern in model-dependent ways. Some models show a reduced Gap: GPT-5.1 narrows from 56.0 to 25.5, Hulu-Med-32B from 77.0 to 43.5, and Gemini~3~Pro from 36.5 to 14.5. In contrast, others become more imbalanced: Claude Sonnet 4.5 increases from 7.5 to 47.0, and LingShu-32B from 35.5 to 92.0. These results suggest that decoupling span detection from significance judgment does not eliminate errors, but redistributes them between Discrimination and Robustness.

\paragraph{Discrimination bias}
Another consistent pattern emerges across both paradigms: nearly all models exhibit a large Gap, with D substantially exceeding R. 
This reflects a systematic tendency to label clinically harmless variations as significant errors, which we term \emph{discrimination bias}. Without explicit supervision on the clinical-significance boundary, models favor sensitivity over selectivity, detecting clinically meaningful errors but failing to reliably separate them from insignificant variation. Since 2-Pass inference only redistributes this bias rather than resolving it, we next explore targeted data augmentation and fine-tuning in
Sections~\ref{sec:data}--\ref{sec:training}.

\begin{figure*}[t!]
\centering
\includegraphics[width=0.85\textwidth]{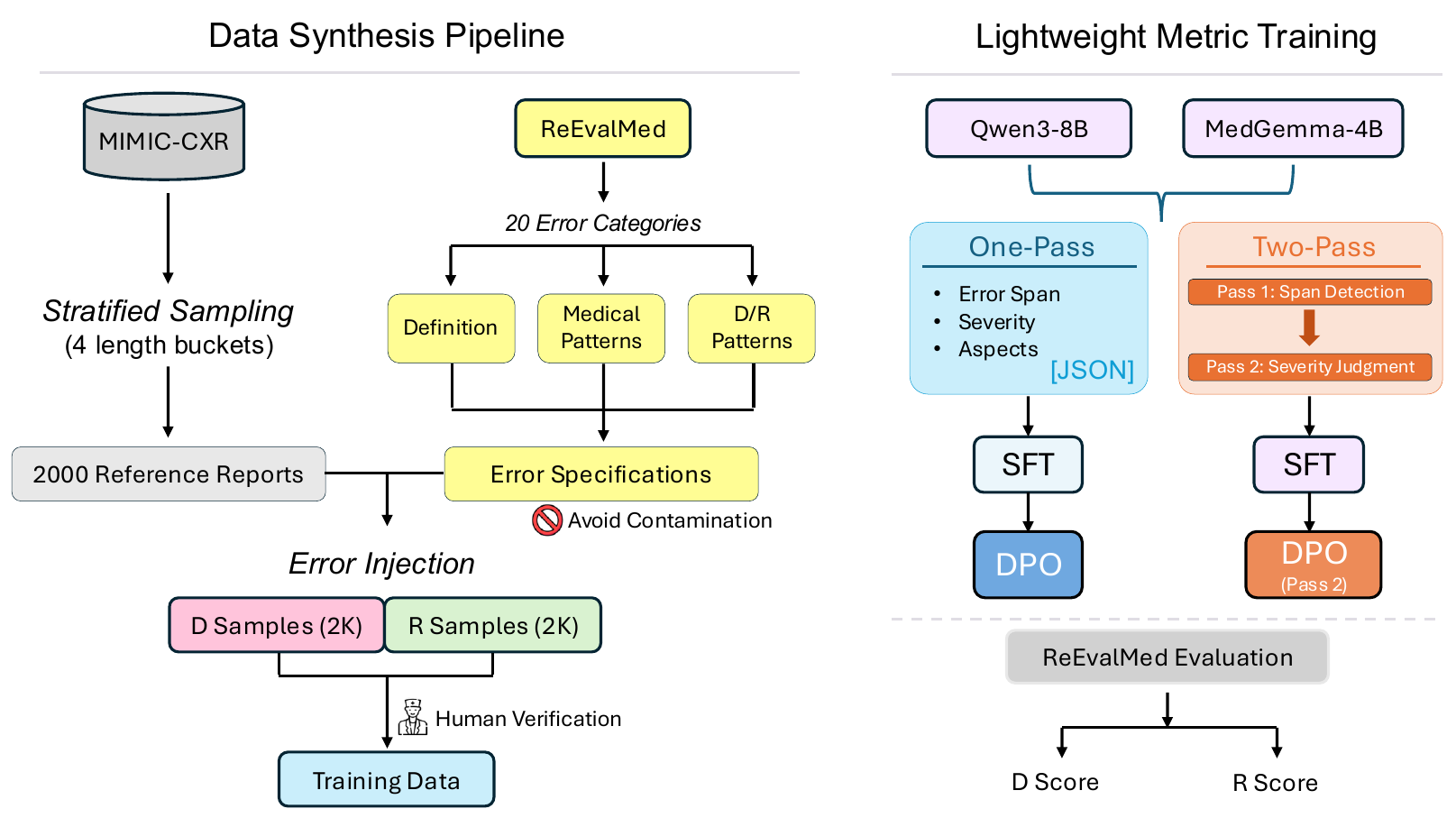}
\caption{Overview of our data synthesis and training pipeline. \textbf{Left:} Reference reports are stratified-sampled from MIMIC-CXR and paired with error specifications for controlled error injection. \textbf{Right:} Base models are fine-tuned through SFT and RL under both One-Pass and Two-Pass evaluation paradigms.}
\label{fig:pipeline}
\end{figure*}

\section{Report Synthesis via Clinical Error Injection}
\label{sec:data}

Section~\ref{sec:benchmark} shows that discrimination bias is pervasive across LLM evaluators, yet no large-scale resource explicitly supervises the boundary between clinically significant errors and harmless report variations. We therefore construct 4,000 REF–TGT medical report pairs with fine-grained annotations across 12 error aspects (omission, fabrication, and inaccuracy) and balanced D–R supervision. This resource is designed to mitigate the bias when training lightweight metrics and to serve as a reusable resource, as shown in Figure~\ref{fig:pipeline} (left).

\subsection{Stratified Sampling}

Error categories naturally differ in their length requirements: \emph{Stylistic Variation} typically requires longer, multi-sentence reports, whereas \emph{Severity Omission} can occur in short single-finding reports. Following stratified sampling practice in machine translation evaluation~\cite{saldias-fuentes-etal-2022-toward}, we sample 2,000 reference reports from the 227K MIMIC-CXR reports~\cite{johnson2019mimic} using four character-length buckets: 0--200, 200--350, 350--500, and $>$500. Category-specific keyword filters derived from the error specifications ensure lexical relevance. The resulting length distribution across the 12 error aspects is shown in Appendix~\ref{app:length_dist}.

\subsection{Error Specifications}

A central challenge is distinguishing clinically significant textual differences from harmless rewrites. 
We address this by prompting Claude Sonnet 4.5 to analyse the ReEvalMed test samples for each error category and distil structured \emph{error specifications}, each containing:
(1)~a precise clinical error definition,
(2)~common medical manifestation patterns in CXR reports,
(3)~discrimination text patterns showing changes
that alter clinical meaning (e.g.\ ``left upper lobe nodule''
$\to$ ``pulmonary nodule''), and
(4)~robustness text patterns showing clinically harmless
reformulations (e.g.\ ``hepatic lesion in segment~7'' $\to$
``right hepatic lobe lesion'').
To avoid potential test-set leakage, we deliberately prohibit direct reuse or paraphrasing of ReEvalMed examples during generation, using the specifications only as abstract category-level guidance for diverse, novel realizations.
We present the error injection prompt
(Appendix~\ref{app:prompt}) and representative examples in Appendix~\ref{app:error_spec}.

\subsection{Error Injection and Report Synthesis}

Each reference report is paired with one Discrimination injection (clinically significant error) and one Robustness injection(clinically harmless variation).\footnote{We generate a single D/R pair per reference to maximize distributional diversity and reduce overfitting to repeated reference patterns.} Across all error categories, with 100~D and 100~R samples per category, the corpus contains \textbf{4,000 (REF, TGT) pairs}. Claude Sonnet 4.5 generates each TGT conditioned on the sampled REF and the corresponding error specification, and each sample is annotated with span-level labels (error spans, type, severity) to support two-pass training.

\subsection{Quality Control via Automatic Sanity Checks and Human Evaluation}

To ensure generation quality, we add automatic sanity checks
(e.g.\ non-empty output, reasonable length ratio) to detect
obviously invalid generations and trigger immediate
A qualified clinician further reviewed 400 randomly sampled pairs from the 4,000 synthesized samples, consisting of 200 Discrimination and 200 Robustness pairs, to assess clinical validity and label quality.

Overall, \textbf{93.0\%} of the reviewed samples passed clinician verification, with 28 of 400 pairs flagged. Manual inspection showed that the flagged cases mainly fell into three categories: 
severity mislabelling (the assigned severity disagrees with
clinical judgement), injection failure (error injection produces
no meaningful change), and aspect mismatch (the injected change
is purely stylistic, deviating from the assigned error type).
Representative flagged cases are provided in Appendix~\ref{app:human_eval} as practical references for constructing clinically grounded datasets. We will also release the complete clinician review report and detailed statistical analysis to support transparency and reproducibility.

\section{Lightweight Metric Training}
\label{sec:training}

\subsection{Training Stages}

We evaluate the synthesized 4K corpus by training a lightweight radiology report metric and testing it on ReEvalMed, measuring both evaluation accuracy and discrimination bias. The corpus is balanced 1:1 between Discrimination and Robustness samples to discourage degenerate always-positive or always-negative predictions. As shown in Figure~\ref{fig:pipeline} (right), both one-pass and two-pass variants follow the same two-stage pipeline: supervised fine-tuning (SFT) for output-format learning, followed by reinforcement learning to align predictions with the clinical significance boundary.

\paragraph{Base models}
We select two lightweight LLMs of different scales:
\textbf{Qwen3-8B}~\cite{yang2025qwen3}, a general-purpose 8B instruction-tuned LLM,
and \textbf{MedGemma-4B}~\cite{sellergren2025medgemma}, a 4B model pre-trained on biomedical corpora.
This allows us to examine whether domain-specific pre-training
complements our clinically grounded data.

\paragraph{SFT}
The SFT stage trains the model to produce the target output format and
acquire basic error detection ability.
For one-pass, the model learns to produce a structured JSON
object containing severity buckets and error spans from a single
REF+TGT input.
For two-pass, we construct separate training samples for Pass~1 and
Pass~2, then fine-tune the same base model on both parts so that it
masters error span extraction (Pass~1: a JSON array of error spans with
clinical aspects) and severity classification (Pass~2: labelling each
span as Critical, Significant, or Insignificant).

\paragraph{RL}
SFT learns the output format and may improve performance,
but does not explicitly optimize the clinical significance boundary.
We apply DPO~\cite{rafailov2023direct} to explicitly optimize this boundary. We apply DPO~\cite{rafailov2023direct} to explicitly optimize this boundary. Using the same 4K corpus, we construct preference pairs by treating the original ground-truth severity label as the \emph{chosen} response and its flipped counterpart (e.g., Significant $\rightarrow$ Insignificant) as the \emph{rejected} response. 
Note that for two-pass, since SFT already learns span extraction reliably, DPO is applied only to Pass~2 severity classification.

\paragraph{Training framework}
All models are trained using LoRA~\cite{hu2022lora} via
LLaMA-Factory~\cite{zheng2024llamafactory}.
Detailed hyperparameters are provided in Appendix~\ref{app:training}.

\subsection{Results}

\begin{table}[t]
\centering
\footnotesize
\setlength{\tabcolsep}{2pt}
\renewcommand{\arraystretch}{1.2}
\begin{tabular}{l!{\vrule}lcccc}
\toprule
\multicolumn{1}{l}{\textbf{Model}} & \textbf{Setting} & \textbf{D\,($\uparrow$)} & \textbf{R\,($\uparrow$)} & \textbf{Avg\,($\uparrow$)} & \textbf{Gap\,($\downarrow$)} \\
\midrule
\multicolumn{6}{l}{\textit{One-Pass}} \\
\hdashline\noalign{\vskip 2pt}
\multirow{3}{*}{Qwen3}
  & Base      & 96.5 & 8.0  & 52.3{\scriptsize\,($\pm$2.3)} & 88.5 \\
  & + SFT     & 90.0 & 19.0 & 54.5{\scriptsize\,($\pm$3.5)} & 71.0 \\
  & + SFT + RL & 86.0 & 39.5 & 62.8{\scriptsize\,($\pm$4.2)} & 46.5 \\
\noalign{\vskip 2pt}
\multirow{3}{*}{MedGemma}
  & Base      & 49.0 & 9.0  & 29.0{\scriptsize\,($\pm$4.0)} & 40.0 \\
  & + SFT     & 60.5 & 93.0 & 76.8{\scriptsize\,($\pm$4.0)} & 32.5 \\
  & + SFT + RL & 75.5 & 80.0 & \textbf{77.8}{\scriptsize\,($\pm$4.2)} & 4.5 \\
\midrule
\multicolumn{6}{l}{\textit{Two-Pass}} \\
\hdashline\noalign{\vskip 2pt}
\multirow{3}{*}{Qwen3}
  & Base      & 95.5 & 9.0  & 52.3{\scriptsize\,($\pm$2.5)} & 86.5 \\
  & + SFT     & 84.0 & 52.0 & 68.0{\scriptsize\,($\pm$4.3)} & 32.0 \\
  & + SFT + RL & 78.5 & 70.5 & 74.5{\scriptsize\,($\pm$4.2)} & 8.0 \\
\noalign{\vskip 2pt}
\multirow{3}{*}{MedGemma}
  & Base      & 67.0 & 67.5 & 67.3{\scriptsize\,($\pm$4.8)} & \textbf{0.5} \\
  & + SFT     & 67.0 & 84.0 & 75.5{\scriptsize\,($\pm$4.2)} & 17.0 \\
  & + SFT + RL & 66.5 & 90.5 & \textbf{78.5}{\scriptsize\,($\pm$4.0)} & 24.0 \\
\bottomrule
\end{tabular}
\renewcommand{\arraystretch}{1}
\caption{Training results on ReEvalMed based on Qwen3-8B~\cite{yang2025qwen3} and MedGemma-4B~\cite{sellergren2025medgemma}. D and R scores are in accuracy (\%). Best results are in \textbf{bold}.}
\label{tab:training}
\end{table}

Table~\ref{tab:training} summarises the training results.

\paragraph{Best results}
After post-training, MedGemma-4B under 1-Pass achieves the
best D--R alignment (D=75.5\%, R=80.0\%, Gap=4.5), while Qwen3-8B
under 2-Pass reaches D=78.5\%, R=70.5\% (Gap=8.0).
Both surpass 32B-scale medical LLMs such as LingShu-32B and
Hulu-Med-32B (Table~\ref{tab:main}), and approach Claude Sonnet 4.5
(D=91.5\%, R=84.0\%) at a fraction of the inference cost.

\paragraph{SFT and RL serve complementary roles}
SFT teaches models to produce structured outputs, but does not explicitly optimize the clinical significance boundary:
Qwen3-8B remains heavily D-biased after SFT
(1-Pass Gap: 88.5$\to$71.0; 2-Pass Gap: 86.5$\to$32.0),
while MedGemma-4B overcorrects toward R-bias under 1-Pass
(D=60.5\%, R=93.0\%).
RL then narrows the Gap from both directions:
for Qwen3-8B it boosts R while preserving D
(2-Pass Gap: 32.0$\to$8.0);
for MedGemma-4B it recovers D from 60.5\% to 75.5\% while
maintaining high R (Gap: 32.5$\to$4.5).
The two stages are therefore complementary: SFT establishes the output format, while RL explicitly optimizes the D--R decision boundary.

\paragraph{1-Pass vs.\ 2-Pass depends on the model}
For Qwen3-8B, 2-Pass substantially outperforms 1-Pass
(Gap 8.0 vs.\ 46.5), as decoupling span detection from severity
judgment helps the model avoid conflating \emph{where} an error is
with \emph{how serious} it is.
For MedGemma-4B, however, 1-Pass yields better alignment
(Gap 4.5 vs.\ 24.0): under 2-Pass, R continues to rise
(84.0$\to$90.5\%) but D stagnates at 66.5\%, amplifying R-bias
with each training stage.
This suggests the the optimal pipeline depends on the base model's inherent D/R bias.

\paragraph{Effectiveness of data and training}
Across both models and paradigms, training on our synthesized data
consistently reduces the Gap, demonstrating the effectiveness of
targeted data augmentation for D--R alignment.
Even with lightweight LoRA training on models as small as 4B,
the resulting metrics achieve competitive performance with
proprietary LLMs, validating both the quality of our synthesized
dataset and the training framework.

\section{Related Work}
\label{sec:related}

\paragraph{Radiology report generation}

Paired image--report datasets such as CheXpert Plus~\cite{chambon2024chexpertplus}, PadChest~\cite{bustos2020padchest}, and the IU X-Ray collection distributed through Open-i~\cite{demner2016preparing} have enabled radiology report generation models, including recent multimodal LLMs such as $\mu^2$LLM~\cite{li2025mu2tokenizer}.
Unlike this generation-focused literature, our work evaluates whether automatic metrics can identify clinically significant errors generated by report generation LLMs while tolerating clinically insignificant report variation.

\paragraph{Radiology report evaluation}
Lexical metrics~\cite{papineni2002bleu,lin2004rouge,banerjee2005meteor} remain common despite limitations for clinical text.
RadEval~\cite{xu2025radeval} unifies lexical, contextual, clinical concept-based, and LLM-based radiology text metrics.
CREPE~\cite{cho2025crepe}, CheXbert~\cite{smit2020combining}, and RadGraph~\cite{jain2021radgraph} provide error counts or structured representations but still reduce quality to scalar scores.
GREEN~\cite{ostmeier2024green}, RGRG~\cite{tanida2023interactive}, and VERT~\cite{bologna2026vert} use LLMs as judges, while CLEAR~\cite{jiang-etal-2025-clear} evaluates CheXpert conditions and fine-grained attributes; none jointly decompose error types and D--R behavior.
RaTEScore~\cite{zhao2024ratescore} and MAIRA~\cite{hyland2023maira} target factual grounding but remain scalar.
ReEvalMed~\cite{li2025reevalmed} introduces D--R evaluation and reveals failures of existing metrics; we build on it with targeted data construction and post-training.

\paragraph{LLM-as-judge}
\citet{zheng2023judging} established the LLM-as-judge paradigm for general
NLP tasks.
In the medical domain, \citet{singhal2023large} demonstrated expert-level
performance on clinical QA\@.
However, LLMs used as report evaluators tend to over-penalise stylistic
variations~\cite{liu2023g}, a failure mode our Robustness dimension
systematically quantifies.

\paragraph{LLM post-training}
Reinforcement learning from human feedback~\cite{ouyang2022training} and DPO~\cite{rafailov2023direct} are standard alignment methods, while LoRA~\cite{hu2022lora} and LLaMA-Factory~\cite{zheng2024llamafactory} make SFT--RL post-training efficient and accessible.
Recent work further simplifies preference optimisation by removing the reference model~\cite{hong2024orpo}.
We use this training stack to build a lightweight D--R-aligned clinical report evaluator.

\section{Conclusion}
\label{sec:conclusion}

We systematically evaluate LLM-based metrics for clinical significance in radiology reports, revealing a consistent \emph{discrimination bias}: LLMs readily detect true clinical errors but struggle to tolerate harmless variations.
To mitigate this, we synthesize 4k report pairs and train lightweight metrics via SFT and RL, sharpening the significance boundary to surpass 32B-scale medical LLMs while remaining competitive with proprietary LLMs.
We also find two-pass inference redistributes rather than eliminates errors between discrimination and robustness.

\section*{Limitations}
The limitations of this study can be summarised as follows:
\begin{itemize}[leftmargin=*,topsep=4pt,itemsep=4pt]
  \item \textbf{Quality control.}
  The most rigorous validation of synthesized report pairs would
  require expert clinician annotation, which is prohibitively
  expensive at scale.
  As a practical compromise, we randomly sampled 400 generated pairs
  for manual verification (Section~\ref{sec:data}), leaving
  full-scale clinician review for future work.

  \item \textbf{Training algorithm.}
  The primary goal of this paper is to explore the proposed
  LLM-based evaluation framework and demonstrate the effectiveness
  of our synthesized data; metric training serves as a validation
  vehicle rather than the core contribution.
  We therefore adopt DPO as a straightforward offline preference
  optimisation method.
  Modern reinforcement learning algorithms (e.g.\ PPO, GRPO)
  typically require large-scale rollouts that exceed our current
  data budget; exploring these and other advanced training pipelines
  is a promising direction for future work.

  \item \textbf{Domain coverage.}
  Our models are trained and evaluated on MIMIC-CXR (English, US
  clinical setting) and Open-i for domain generalization; out-of-distribution generalisation to other
  languages, imaging modalities, or clinical environments remains
  to be studied.

  \item \textbf{Inference cost.}
  The two-pass approach doubles inference cost relative to one-pass. In high-throughput settings, one-pass DPO may therefore provide a more favorable trade-off between efficiency and D/R alignment.

\end{itemize}

\section*{Ethics Statement}
We take ethical considerations seriously and strictly adhere to the ACL Code of Ethics.
All data used in this study are derived from publicly available research datasets.
For MIMIC-CXR, we complied with the PhysioNet credentialing process and data-use agreement requirements, including completion of the required human-subjects research training and access under the approved data-use terms.
No additional protected patient information was collected or used beyond the de-identified data available through the authorized dataset access.
A clinician co-author participated in the human verification of synthesized data (Section~\ref{sec:data}) to ensure clinical validity.
Our proposed approaches and metrics do not include statements that induce the model to generate harmful information; the evaluation framework focuses solely on assessing the clinical significance of textual differences in radiology reports.
We ensure that the findings and conclusions of this paper are reported accurately and objectively.

\bibliography{references}

\appendix

\section{Model and Training Hyperparameters}
\label{app:training}

Table~\ref{tab:hyperparams} lists the model configuration and training
hyperparameters used in all experiments.
We fine-tune two base models (Qwen3-8B and
MedGemma-4B) using LoRA~\cite{hu2022lora} via
LLaMA-Factory~\cite{zheng2024llamafactory}.
SFT and DPO share the same LoRA and optimiser settings; DPO-specific
parameters (preference construction) are listed separately.

\begin{table}[h]
\centering
\small
\renewcommand{\arraystretch}{1.2}
\begin{tabular}{ll}
\toprule
\textbf{Hyperparameter} & \textbf{Value} \\
\midrule
\multicolumn{2}{l}{\textit{Model}} \\
\hdashline\noalign{\vskip 2pt}
Base model 1           & Qwen/Qwen3-8B (8B) \\
Base model 2           & google/medgemma-4b-it (4B) \\
\midrule
\multicolumn{2}{l}{\textit{LoRA}} \\
\hdashline\noalign{\vskip 2pt}
LoRA rank              & 8 \\
LoRA alpha             & 16 \\
\midrule
\multicolumn{2}{l}{\textit{Training}} \\
\hdashline\noalign{\vskip 2pt}
Learning rate          & 1e-4 \\
Epochs                 & 3 \\
Scheduler              & cosine \\
Precision              & bfloat16 \\
Gradient accumulation  & 8 \\
Max sequence length    & 2048 \\
\midrule
\multicolumn{2}{l}{\textit{DPO}} \\
\hdashline\noalign{\vskip 2pt}
$\beta$                & 0.1 \\
Sampling temperature   & 1.0 \\
Samples per input      & 2 \\
\bottomrule
\end{tabular}
\renewcommand{\arraystretch}{1}
\caption{Model and training hyperparameters for all SFT and DPO
experiments. LoRA and optimiser settings are shared across stages;
DPO-specific parameters govern preference pair construction via
rejection sampling.}
\label{tab:hyperparams}
\end{table}

\section{Additional Analysis}
\label{app:analysis}

\subsection{Generalization to Open-i}
\label{app:openi_generalization}

\begin{table*}[t]
\centering
\footnotesize
\renewcommand{\arraystretch}{1.2}
\setlength{\tabcolsep}{5pt}
\begin{tabular}{clcccc}
\toprule
\textbf{Type} & \textbf{Model} & \textbf{D\,($\uparrow$)} & \textbf{R\,($\uparrow$)} & \textbf{Avg\,($\uparrow$)} & \textbf{Gap\,($\downarrow$)} \\
\midrule
\multirow{3}{*}{Baseline}
 & RadFact       & \textbf{100.0} & \textbf{100.0} & \textbf{100.0} & \textbf{0.0} \\
 & FineRadScore  & \textbf{100.0} & \textbf{100.0} & \textbf{100.0} & \textbf{0.0} \\
 & CRIMSON       & 92.0 & \textbf{100.0} & 96.0 & 8.0 \\
\midrule
\multirow{1}{*}{Proprietary}
 & Claude Sonnet 4.5 & \textbf{100.0} & \textbf{100.0} & \textbf{100.0} & \textbf{0.0} \\
\midrule
\multirow{4}{*}{Open-source}
 & Qwen3-8B                 & \textbf{100.0} & 76.0 & 88.0 & 24.0 \\
 & Qwen3-8B + SFT + RL      & \textbf{100.0} & 96.0 & 98.0 & 4.0 \\
 & MedGemma-4B              & \textbf{100.0} & 92.0 & 96.0 & 8.0 \\
 & MedGemma-4B + SFT + RL   & \textbf{100.0} & \textbf{100.0} & \textbf{100.0} & \textbf{0.0} \\
\bottomrule
\end{tabular}
\caption{Comparison of evaluators on the Open-i 50-sample benchmark. D and R scores are in accuracy (\%). Avg is the mean of D and R, and Gap is the absolute difference between D and R.}
\label{tab:openi_50}
\renewcommand{\arraystretch}{1}
\end{table*}

We further constructed a small Open-i benchmark to test evaluator performance on an external chest X-ray dataset. 
Following the ReEvalMed pipeline, we sampled 50 cases from Open-i and applied controlled data augmentation to create 25 significant-error samples for discrimination and 25 insignificant-error samples for robustness. 
To avoid potential same-model bias in the generation of evaluation samples, we used GPT-5.5 for this augmentation process; GPT-5.5 is not included among the evaluator models assessed in this paper. 
All augmented samples were reviewed by clinicians to ensure that the injected errors were correct, clinically meaningful when intended, and free from duplicates or ambiguous cases.

As shown in Table~\ref{tab:openi_50}, most methods achieve strong results on Open-i. One likely reason is that Open-i reports are generally simple and contain relatively short sentences. This makes both significant-error detection and robustness to minor wording changes easier than in ReEvalMed. However, the results still reveal differences among evaluators. Qwen3-8B shows strong discrimination but weaker robustness, indicating a tendency to over-detect minor changes as significant errors. After SFT and RL, Qwen3-8B achieves better balance. MedGemma-4B performs well on the base model and reaches perfect performance after SFT and RL.

\subsection{Error Aspect Analysis}
\label{app:error_aspect_analysis}

Beyond the overall D--R gap, we further ask \emph{which} types of
clinical perturbations are responsible for evaluator errors.
Specifically, we conduct an aspect-level confusion analysis by comparing
the ground-truth error aspect assigned during dataset construction with
the aspect predicted by each LLM evaluator. This analysis allows us to
move beyond aggregate discrimination and robustness accuracy and inspect
whether models confuse, over-predict, or under-recognize particular
clinical error categories, such as negation, location, uncertainty, or
stylistic variation.

\paragraph{LLM Evaluators} Figure~\ref{fig:app_sankey_all}
visualizes the aspect-level flows for eight LLM evaluators, with one-pass
predictions on the left, ground-truth aspects in the middle, and two-pass
predictions on the right.  A common pattern is that model predictions collapse
toward a small set of broad clinical aspects, especially Description, Location,
Severity, and Comparison/Progression, while fine-grained categories such as
Modality, Noise, Uncertainty, Terminology, and Stylistic Variation are often
under-recognized.  For example, GPT-5.1 and Claude Sonnet 4.5 both over-predict
Description under two-pass inference (60$\to$113 and 60$\to$96, respectively),
whereas Qwen3-8B and MedGemma-4B show even stronger concentration, mapping many
cases to Description or Location.  Qwen3-Max is comparatively better calibrated,
but still shifts substantially toward Location in the two-pass setting
(60$\to$99).

The Sankey diagrams also show that two-pass inference does not simply ``fix''
one-pass errors; rather, it redistributes them.  In several models, the second
pass reduces some over-predicted categories but introduces new failure modes,
including large Unclassified flows for Gemini 3 Pro and LingShu-32B, suggesting
that span extraction and severity judging can propagate formatting or
classification errors.  Across models, explicit factual aspects such as Location,
Severity, and Negation are more frequently detected, while subtle imaging- or
wording-specific aspects such as Modality, Noise, and Terminology remain
difficult.  This aspect-level evidence supports our central finding that current
LLM evaluators tend to over-detect discrepancies but remain under-calibrated in
distinguishing clinically meaningful errors from harmless variations.

\begin{figure*}[p]
\centering
\begin{minipage}[t]{0.49\textwidth}
  \centering
  \includegraphics[width=\linewidth,height=0.88\textheight,keepaspectratio]{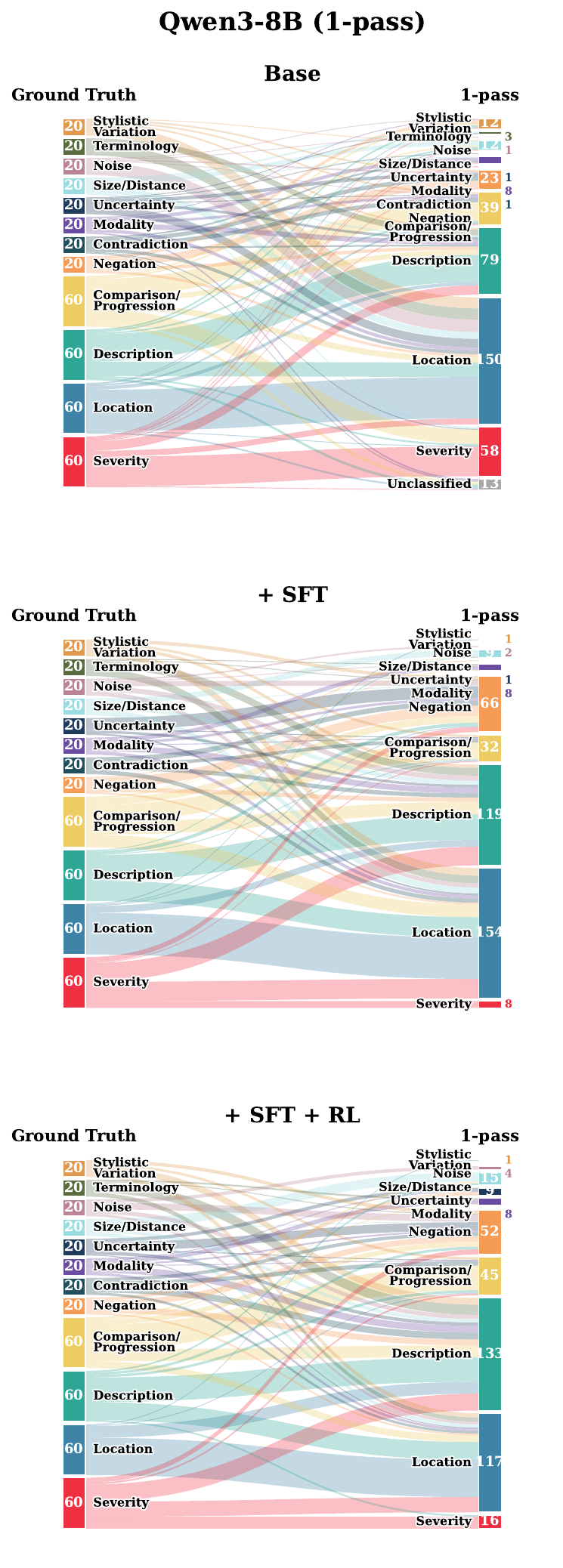}\\[-2pt]
  \textbf{(a) Qwen3-8B, 1-pass}
\end{minipage}
\hfill
\begin{minipage}[t]{0.49\textwidth}
  \centering
  \includegraphics[width=\linewidth,height=0.88\textheight,keepaspectratio]{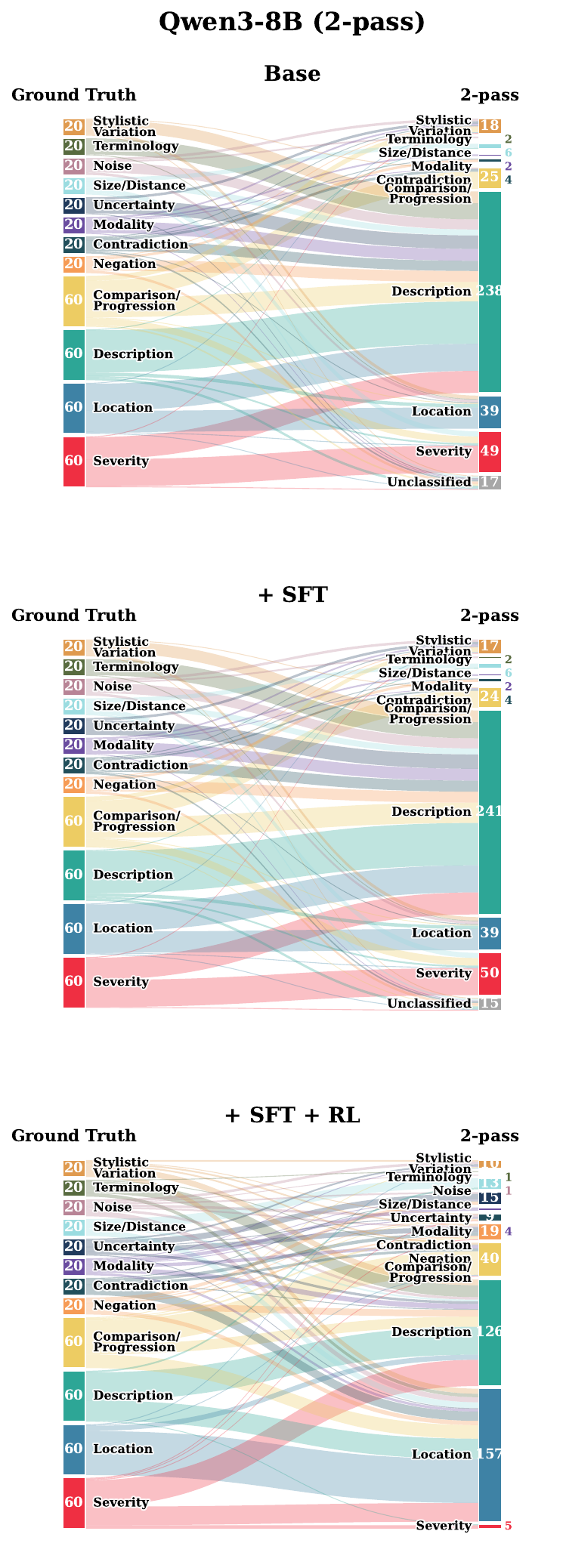}\\[-2pt]
  \textbf{(b) Qwen3-8B, 2-pass}
\end{minipage}
\caption{
Aspect-level Sankey comparison across training stages for Qwen3-8B.
Each panel visualizes the flow from ground-truth error aspects to model-predicted aspects under Base, SFT, and SFT+RL settings.
}
\label{fig:app_qwen_training_sankey}
\end{figure*}

\paragraph{Training Comparison.}
Figure~\ref{fig:app_qwen_training_sankey} shows that Qwen3-8B exhibits strong aspect-level bias before post-training. In the one-pass setting, the base model over-predicts broad and visually salient categories such as Location and Description, while rarely assigning fine-grained aspects such as Noise, Terminology, or Uncertainty. SFT improves the output distribution only partially: it increases detection of Negation but still collapses many cases into Location and Description. RL produces a more redistributed one-pass prediction pattern, reducing the extreme Location bias and increasing flows into Comparison/Progression, Negation, and other clinically meaningful aspects. In the two-pass setting, however, Base and SFT both heavily collapse into Description, suggesting that simply teaching the two-stage format does not resolve aspect confusion. The RL stage substantially changes this behavior by shifting mass away from Description and toward Location and Comparison/Progression, which is consistent with the improved D--R balance reported in Table~\ref{tab:training}.

\begin{figure*}[p]
\centering
\begin{minipage}[t]{0.49\textwidth}
  \centering
  \includegraphics[width=\linewidth,height=0.88\textheight,keepaspectratio]{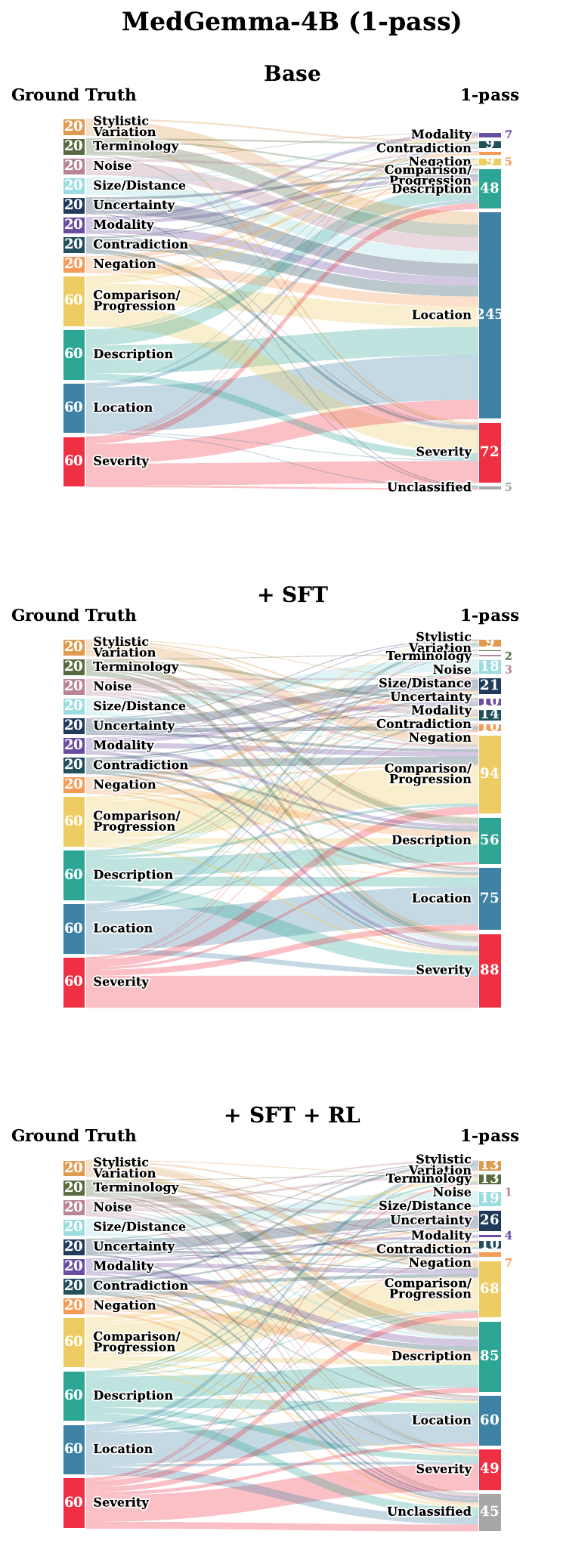}\\[-2pt]
  \textbf{(a) MedGemma-4B, 1-pass}
\end{minipage}
\hfill
\begin{minipage}[t]{0.49\textwidth}
  \centering
  \includegraphics[width=\linewidth,height=0.88\textheight,keepaspectratio]{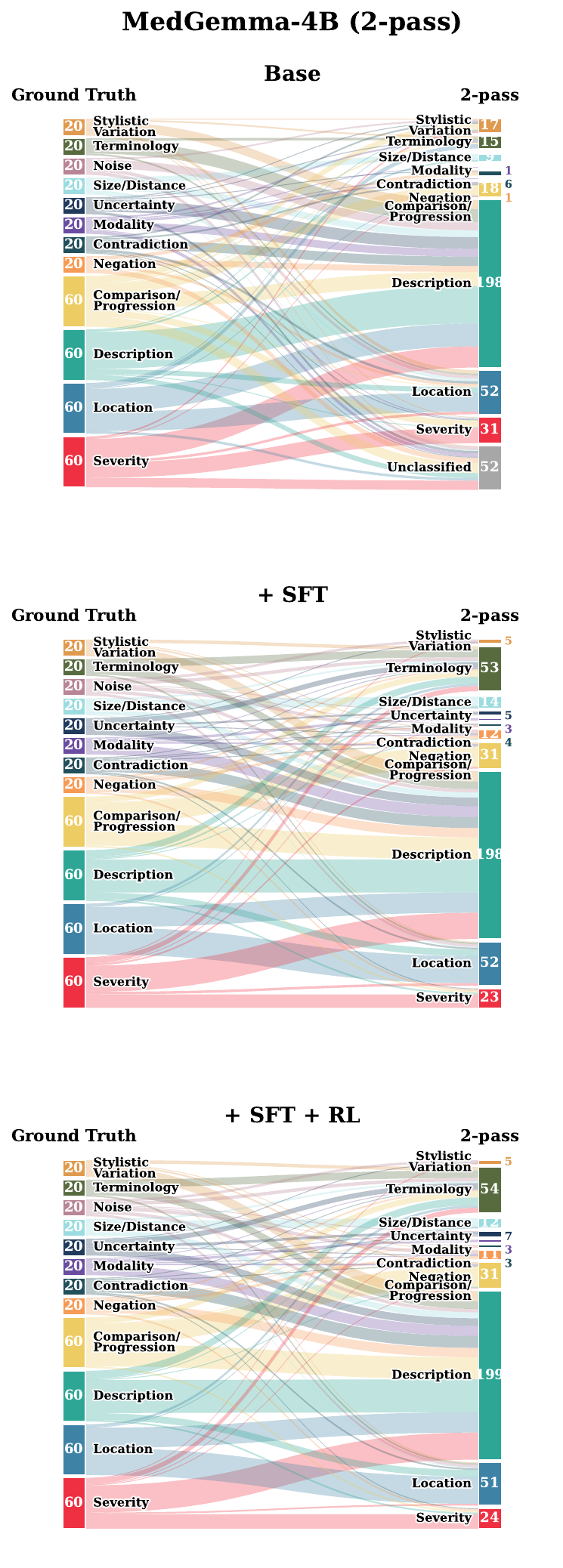}\\[-2pt]
  \textbf{(b) MedGemma-4B, 2-pass}
\end{minipage}
\caption{
Aspect-level Sankey comparison across training stages for MedGemma-4B.
Each panel visualizes the flow from ground-truth error aspects to model-predicted aspects under Base, SFT, and SFT+RL settings.
}
\label{fig:app_medgemma_training_sankey}
\end{figure*}

Figure~\ref{fig:app_medgemma_training_sankey} shows a different pattern for MedGemma-4B. In one-pass inference, the base model is dominated by Location predictions, indicating a strong tendency to interpret diverse report discrepancies as anatomical-site errors. After SFT, the distribution becomes much broader, with substantial flows into Comparison/Progression, Severity, Description, Uncertainty, and Size/Distance. Adding RL further smooths the prediction distribution and reduces the Location collapse, which helps explain why MedGemma-4B achieves the best one-pass D--R alignment after SFT+RL. In contrast, the two-pass MedGemma variants remain highly concentrated on Description, with SFT and RL additionally increasing Terminology predictions. This suggests that MedGemma benefits strongly from post-training in the one-pass setting, whereas its two-pass behavior is more constrained by the aspect labels produced during span extraction.

\subsection{Statistical Analysis} \label{app:stat_analysis}

\paragraph{Confidence intervals.}
We report 95\% bootstrap confidence intervals (CIs) for all evaluation scores in Table~\ref{tab:appbaseline}, Table~\ref{tab:app_main} and Table~\ref{tab:app_training}. For each metric or model, we resample test cases with replacement and recompute Discrimination (D), Robustness (R), Avg, and Gap over 10,000 bootstrap replicates. We report the half-width of the 95\% percentile bootstrap CI in parentheses. All CIs are reported in percentage points.

Across these tables, the 95\% bootstrap CIs are generally moderate, with Avg half-widths mostly around 3--5 percentage points. The large gaps between conventional similarity metrics and the best LLM-based evaluators are therefore unlikely to be explained by sampling noise alone. By contrast, several top-performing LLM-based metrics and models have overlapping CIs, so their small numerical differences should be interpreted cautiously rather than as definitive rankings.

\begin{figure*}[t!]
  \centering
  \includegraphics[width=1.5\columnwidth]{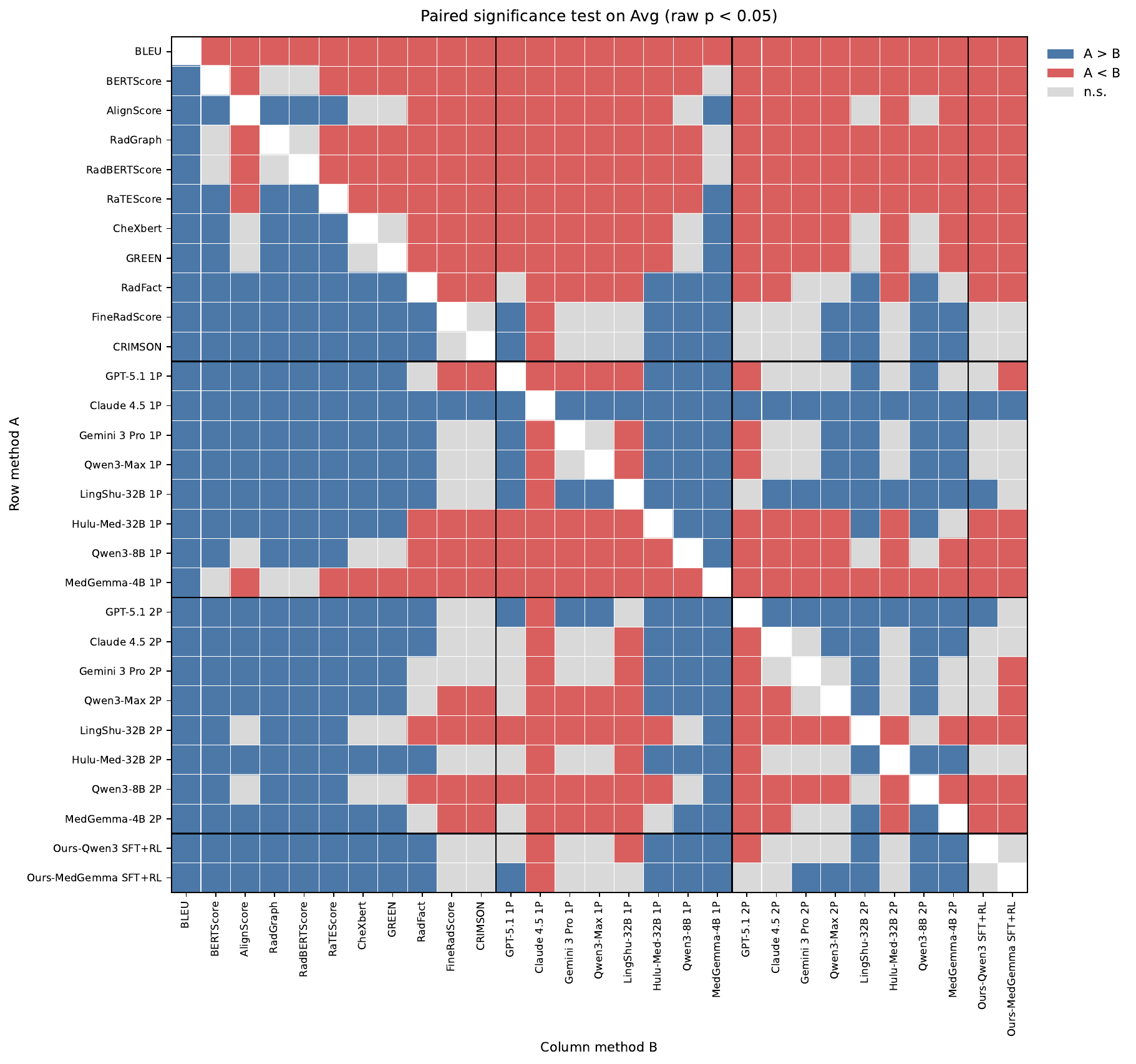}
  \caption{Pairwise paired significance tests on Avg across metrics and evaluators. Each cell compares the row method against the column method using a paired randomization test over matched test cases. Blue indicates that the row method significantly outperforms the column method, red indicates that it performs significantly worse, and gray indicates no significant difference after Holm correction ($p \geq 0.05$).}
  \label{fig:app-paired-significance}
\end{figure*}

\paragraph{Pairwise Significance Test.} We further conduct pairwise paired significance tests to compare metrics and evaluators on Avg in Figure~\ref{fig:app-paired-significance}. For each pair of methods, we compute the per-example correctness difference on the same discrimination and robustness test cases, and test whether the paired Avg difference is significantly different from zero using a paired randomization test. Since all methods are evaluated on the same test instances, this paired test controls for example-level difficulty and is more appropriate than an independent-sample comparison.

The significance matrix shows that broad differences between weak similarity-based metrics and stronger LLM-based evaluators are statistically significant after Holm correction. However, several top-performing methods have no significant differences from each other, including CRIMSON, FineRadScore, and our MedGemma SFT+RL evaluator. This suggests that their small numerical Avg differences should be interpreted cautiously, while the improvements from trained evaluators over their base counterparts are statistically reliable.

\begin{table*}[t]
\centering
\small
\renewcommand{\arraystretch}{1.3}
\setlength{\tabcolsep}{7pt}
\begin{tabular}{cccccc}
\toprule
\multirow{2}{*}{\textbf{Type}} & \multirow{2}{*}{\textbf{Metric}} & \multicolumn{4}{c}{\textbf{Maximin Threshold}} \\
\cmidrule(lr){3-6}
 &  & \textbf{D\,($\uparrow$)} & \textbf{R\,($\uparrow$)} & \textbf{Avg\,($\uparrow$)} & \textbf{Gap\,($\downarrow$)} \\
\midrule
\multirow{3}{*}{NLP}
 & BLEU       & 19.0{\scriptsize\,($\pm$5.5)} & 19.0{\scriptsize\,($\pm$5.5)} & 19.0{\scriptsize\,($\pm$4.0)} & 0.0{\scriptsize\,($\pm$8.0)} \\
 & BERTScore  & 24.0{\scriptsize\,($\pm$6.0)} & 24.0{\scriptsize\,($\pm$6.0)} & 24.0{\scriptsize\,($\pm$4.2)} & 0.0{\scriptsize\,($\pm$8.5)} \\
 & AlignScore & 51.0{\scriptsize\,($\pm$7.0)} & 51.0{\scriptsize\,($\pm$7.0)} & 51.0{\scriptsize\,($\pm$5.0)} & 0.0{\scriptsize\,($\pm$10.5)} \\
\midrule
\multirow{4}{*}{Med.}
 & RadGraph   & 28.5{\scriptsize\,($\pm$6.5)} & 28.5{\scriptsize\,($\pm$6.0)} & 28.5{\scriptsize\,($\pm$4.5)} & 0.0{\scriptsize\,($\pm$10.0)} \\
 & RadBERTScore & 26.5{\scriptsize\,($\pm$6.0)} & 26.5{\scriptsize\,($\pm$6.5)} & 26.5{\scriptsize\,($\pm$4.2)} & 0.0{\scriptsize\,($\pm$10.0)} \\
 & RaTEScore  & 35.5{\scriptsize\,($\pm$7.0)} & 35.5{\scriptsize\,($\pm$7.0)} & 35.5{\scriptsize\,($\pm$4.8)} & 0.0{\scriptsize\,($\pm$10.0)} \\
 & CheXbert   & 47.0{\scriptsize\,($\pm$7.0)} & 49.0{\scriptsize\,($\pm$7.0)} & 48.0{\scriptsize\,($\pm$5.0)} & 2.0{\scriptsize\,($\pm$10.0)} \\
\midrule
\multirow{4}{*}{LLM}
 & GREEN      & 53.5{\scriptsize\,($\pm$7.0)} & 41.5{\scriptsize\,($\pm$7.0)} & 47.5{\scriptsize\,($\pm$4.8)} & 12.0{\scriptsize\,($\pm$10.0)} \\
 & RadFact    & 80.0{\scriptsize\,($\pm$5.5)} & 56.0{\scriptsize\,($\pm$7.0)} & 68.0{\scriptsize\,($\pm$4.5)} & 24.0{\scriptsize\,($\pm$9.5)} \\
 & FineRadScore & \textbf{86.5}{\scriptsize\,($\pm$5.0)} & 68.5{\scriptsize\,($\pm$6.5)} & 77.5{\scriptsize\,($\pm$4.0)} & 18.0{\scriptsize\,($\pm$8.0)} \\
 & CRIMSON    & 75.0{\scriptsize\,($\pm$6.0)} & \textbf{80.5}{\scriptsize\,($\pm$5.5)} & \textbf{77.8}{\scriptsize\,($\pm$4.2)} & 5.5{\scriptsize\,($\pm$8.0)} \\
\bottomrule
\end{tabular}
\caption{Comparison of score-based metrics on ReEvalMed. With Numbers in parentheses denote the half-width of 95\% bootstrap CIs.}
\label{tab:appbaseline}
\renewcommand{\arraystretch}{1.1}
\end{table*}

\begin{table*}[t]
\centering
\footnotesize
\renewcommand{\arraystretch}{1.2}
\setlength{\tabcolsep}{2pt}
\begin{tabular}{cccccccccc}
\toprule
\multirow{2}{*}{\textbf{Type}} & \multirow{2}{*}{\textbf{Model}} & \multicolumn{4}{c}{\textbf{1-Pass}} & \multicolumn{4}{c}{\textbf{2-Pass}} \\
\cmidrule(lr){3-6}\cmidrule(lr){7-10}
 &  & \textbf{D\,($\uparrow$)} & \textbf{R\,($\uparrow$)} & \textbf{Avg\,($\uparrow$)} & \textbf{Gap\,($\downarrow$)} & \textbf{D\,($\uparrow$)} & \textbf{R\,($\uparrow$)} & \textbf{Avg\,($\uparrow$)} & \textbf{Gap\,($\downarrow$)} \\
\midrule
\multirow{3}{*}{Proprietary}
 & GPT-5.1           & 98.5{\scriptsize\,($\pm$2.0)} & 42.5{\scriptsize\,($\pm$7.0)} & 70.5{\scriptsize\,($\pm$3.5)} & 56.0{\scriptsize\,($\pm$7.0)} & 95.0{\scriptsize\,($\pm$3.0)} & 69.5{\scriptsize\,($\pm$6.5)} & \textbf{82.3}{\scriptsize\,($\pm$3.5)} & 25.5{\scriptsize\,($\pm$7.0)} \\
 & Claude Sonnet 4.5 & 91.5{\scriptsize\,($\pm$4.0)} & \textbf{84.0}{\scriptsize\,($\pm$5.5)} & \textbf{87.8}{\scriptsize\,($\pm$3.2)} & \textbf{7.5}{\scriptsize\,($\pm$6.5)} & \textbf{98.0}{\scriptsize\,($\pm$2.0)} & 51.0{\scriptsize\,($\pm$7.0)} & 74.5{\scriptsize\,($\pm$3.5)} & 47.0{\scriptsize\,($\pm$7.0)} \\
 & Gemini 3 Pro      & 94.0{\scriptsize\,($\pm$3.5)} & 57.5{\scriptsize\,($\pm$6.5)} & 75.8{\scriptsize\,($\pm$3.8)} & 36.5{\scriptsize\,($\pm$7.5)} & 65.0{\scriptsize\,($\pm$6.5)} & \textbf{79.5}{\scriptsize\,($\pm$6.0)} & 72.3{\scriptsize\,($\pm$4.2)} & 14.5{\scriptsize\,($\pm$9.0)} \\
\midrule
\multirow{5}{*}{Open-source}
 & Qwen3-Max         & 97.0{\scriptsize\,($\pm$2.5)} & 53.5{\scriptsize\,($\pm$7.0)} & 75.3{\scriptsize\,($\pm$3.8)} & 43.5{\scriptsize\,($\pm$7.5)} & 96.5{\scriptsize\,($\pm$2.5)} & 45.0{\scriptsize\,($\pm$7.0)} & 70.8{\scriptsize\,($\pm$3.8)} & 51.5{\scriptsize\,($\pm$7.5)} \\
 & LingShu-32B       & 99.5{\scriptsize\,($\pm$1.0)} & 64.0{\scriptsize\,($\pm$6.5)} & 81.8{\scriptsize\,($\pm$3.3)} & 35.5{\scriptsize\,($\pm$6.5)} & 97.5{\scriptsize\,($\pm$2.5)} & 5.5{\scriptsize\,($\pm$3.5)} & 51.5{\scriptsize\,($\pm$2.0)} & 92.0{\scriptsize\,($\pm$4.0)} \\
 & Hulu-Med-32B      & \textbf{100.0}{\scriptsize\,($\pm$0.0)} & 23.0{\scriptsize\,($\pm$6.0)} & 61.5{\scriptsize\,($\pm$3.0)} & 77.0{\scriptsize\,($\pm$6.0)} & 96.0{\scriptsize\,($\pm$3.0)} & 52.5{\scriptsize\,($\pm$7.0)} & 74.3{\scriptsize\,($\pm$3.8)} & 43.5{\scriptsize\,($\pm$7.5)} \\
 & Qwen3-8B          & 96.5{\scriptsize\,($\pm$2.5)} & 8.0{\scriptsize\,($\pm$4.0)} & 52.3{\scriptsize\,($\pm$2.3)} & 88.5{\scriptsize\,($\pm$5.0)} & 95.5{\scriptsize\,($\pm$3.0)} & 9.0{\scriptsize\,($\pm$4.0)} & 52.3{\scriptsize\,($\pm$2.5)} & 86.5{\scriptsize\,($\pm$5.0)} \\
 & MedGemma-4B       & 49.0{\scriptsize\,($\pm$7.0)} & 9.0{\scriptsize\,($\pm$4.0)} & 29.0{\scriptsize\,($\pm$4.0)} & 40.0{\scriptsize\,($\pm$8.0)} & 67.0{\scriptsize\,($\pm$6.5)} & 67.5{\scriptsize\,($\pm$6.5)} & 67.3{\scriptsize\,($\pm$4.8)} & \textbf{0.5}{\scriptsize\,($\pm$10.0)} \\
\bottomrule
\end{tabular}
\caption{Comparison of LLM-as-evaluator on ReEvalMed benchmark. D and R scores are in accuracy (\%). Numbers in parentheses denote the half-width of 95\% bootstrap CIs. Best results are in \textbf{bold}.}
\label{tab:app_main}
\renewcommand{\arraystretch}{1.1}
\end{table*}

\begin{table*}[ht]
\centering
\footnotesize
\setlength{\tabcolsep}{5pt}
\renewcommand{\arraystretch}{1.2}
\begin{tabular}{l!{\vrule}lcccc}
\toprule
\multicolumn{1}{l}{\textbf{Model}} & \textbf{Setting} & \textbf{D\,($\uparrow$)} & \textbf{R\,($\uparrow$)} & \textbf{Avg\,($\uparrow$)} & \textbf{Gap\,($\downarrow$)} \\
\midrule
\multicolumn{6}{l}{\textit{One-Pass}} \\
\hdashline\noalign{\vskip 2pt}
\multirow{3}{*}{Qwen3}
  & Base      & 96.5{\scriptsize\,($\pm$3.0)} & 8.0{\scriptsize\,($\pm$4.0)}  & 52.3{\scriptsize\,($\pm$2.3)} & 88.5{\scriptsize\,($\pm$4.5)} \\
  & + SFT     & 90.0{\scriptsize\,($\pm$4.5)} & 19.0{\scriptsize\,($\pm$5.5)} & 54.5{\scriptsize\,($\pm$3.5)} & 71.0{\scriptsize\,($\pm$7.0)} \\
  & + SFT + RL & 86.0{\scriptsize\,($\pm$5.0)} & 39.5{\scriptsize\,($\pm$7.0)} & 62.8{\scriptsize\,($\pm$4.2)} & 46.5{\scriptsize\,($\pm$8.5)} \\
\noalign{\vskip 2pt}
\multirow{3}{*}{MedGemma}
  & Base      & 49.0{\scriptsize\,($\pm$7.0)} & 9.0{\scriptsize\,($\pm$4.0)}  & 29.0{\scriptsize\,($\pm$4.0)} & 40.0{\scriptsize\,($\pm$8.0)} \\
  & + SFT     & 60.5{\scriptsize\,($\pm$7.0)} & 93.0{\scriptsize\,($\pm$3.5)} & 76.8{\scriptsize\,($\pm$4.0)} & 32.5{\scriptsize\,($\pm$7.5)} \\
  & + SFT + RL & 75.5{\scriptsize\,($\pm$6.0)} & 80.0{\scriptsize\,($\pm$5.5)} & \textbf{77.8}{\scriptsize\,($\pm$4.2)} & 4.5{\scriptsize\,($\pm$8.0)} \\
\midrule
\multicolumn{6}{l}{\textit{Two-Pass}} \\
\hdashline\noalign{\vskip 2pt}
\multirow{3}{*}{Qwen3}
  & Base      & 95.5{\scriptsize\,($\pm$3.0)} & 9.0{\scriptsize\,($\pm$4.0)}  & 52.3{\scriptsize\,($\pm$2.5)} & 86.5{\scriptsize\,($\pm$5.0)} \\
  & + SFT     & 84.0{\scriptsize\,($\pm$5.0)} & 52.0{\scriptsize\,($\pm$7.0)} & 68.0{\scriptsize\,($\pm$4.3)} & 32.0{\scriptsize\,($\pm$8.5)} \\
  & + SFT + RL & 78.5{\scriptsize\,($\pm$6.0)} & 70.5{\scriptsize\,($\pm$6.5)} & 74.5{\scriptsize\,($\pm$4.2)} & 8.0{\scriptsize\,($\pm$8.5)} \\
\noalign{\vskip 2pt}
\multirow{3}{*}{MedGemma}
  & Base      & 67.0{\scriptsize\,($\pm$6.5)} & 67.5{\scriptsize\,($\pm$6.5)} & 67.3{\scriptsize\,($\pm$4.8)} & \textbf{0.5}{\scriptsize\,($\pm$10.0)} \\
  & + SFT     & 67.0{\scriptsize\,($\pm$6.5)} & 84.0{\scriptsize\,($\pm$5.0)} & 75.5{\scriptsize\,($\pm$4.2)} & 17.0{\scriptsize\,($\pm$8.5)} \\
  & + SFT + RL & 66.5{\scriptsize\,($\pm$6.5)} & 90.5{\scriptsize\,($\pm$4.5)} & \textbf{78.5}{\scriptsize\,($\pm$4.0)} & 24.0{\scriptsize\,($\pm$7.5)} \\
\bottomrule
\end{tabular}
\renewcommand{\arraystretch}{1}
\caption{Training results on ReEvalMed based on Qwen3-8B~\cite{yang2025qwen3} and MedGemma-4B~\cite{sellergren2025medgemma}. D and R scores are in accuracy (\%). Numbers in parentheses denote the half-width of 95\% bootstrap CIs. Best results are in \textbf{bold}.}
\label{tab:app_training}
\end{table*}

\begin{figure*}[p]
\centering
\setlength{\tabcolsep}{2pt}
\renewcommand{\arraystretch}{0.08}

\begin{tabular}{cc}
\includegraphics[width=0.47\textwidth]{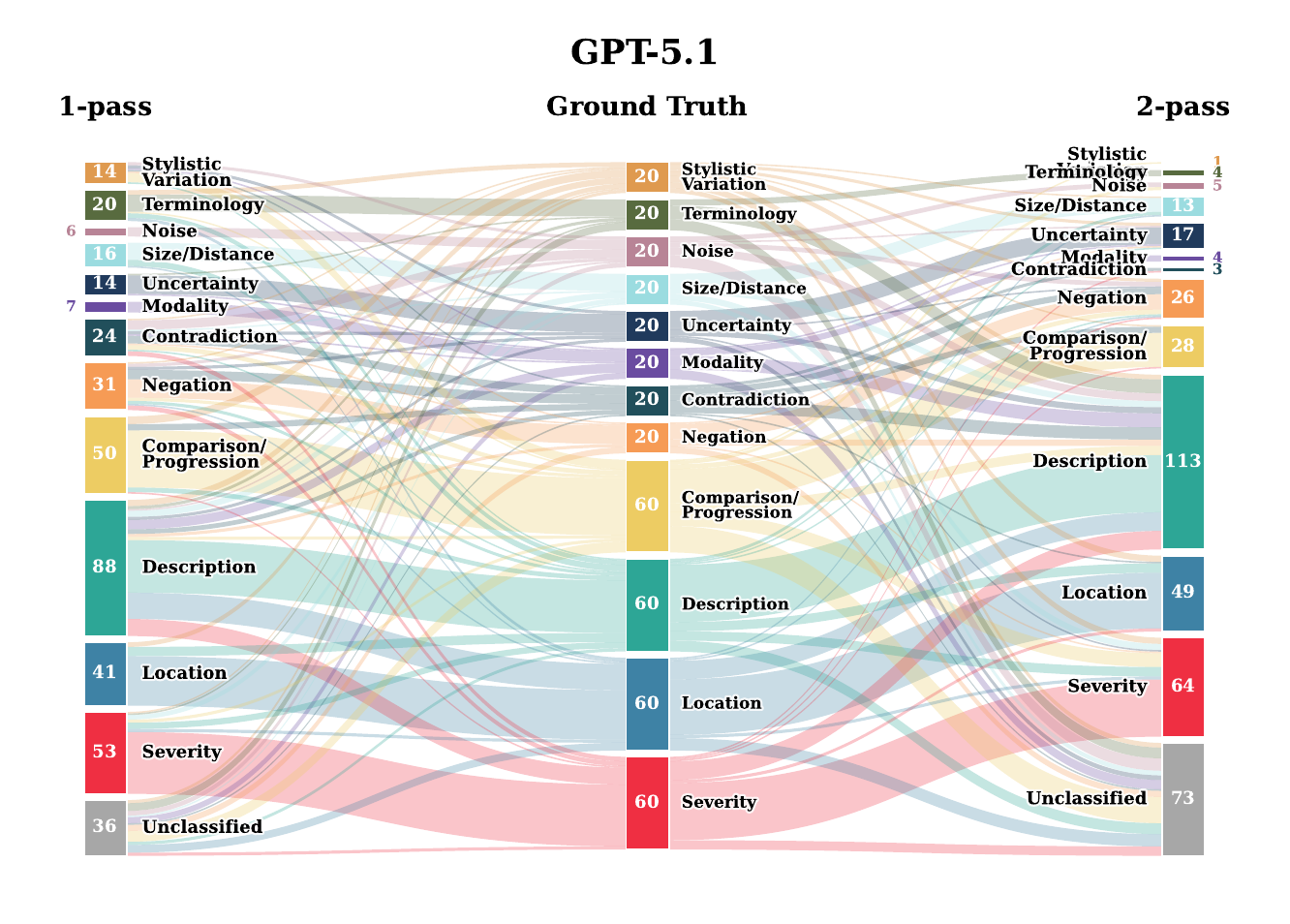} &
\includegraphics[width=0.47\textwidth]{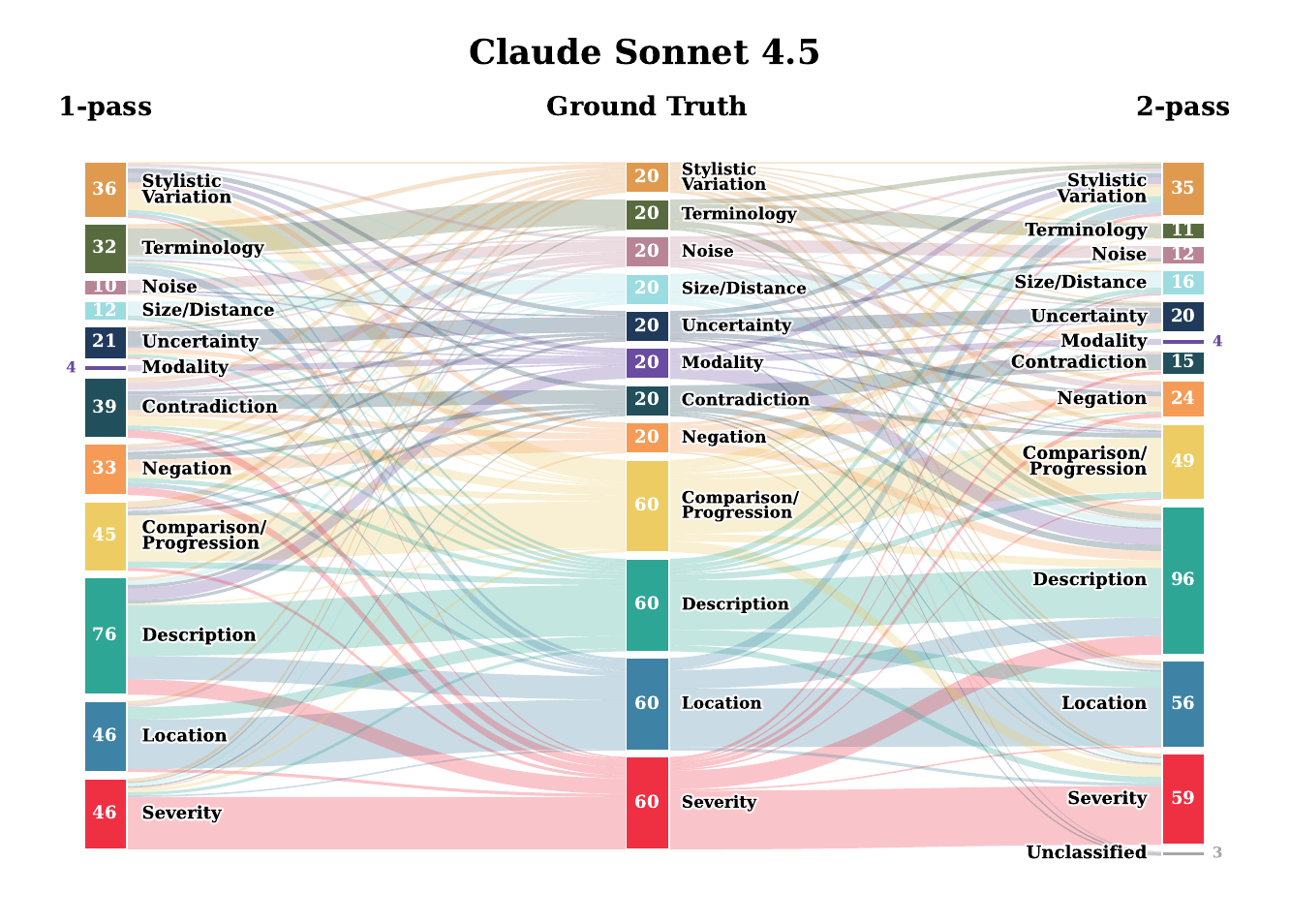} \\[-0.4em]

\includegraphics[width=0.47\textwidth]{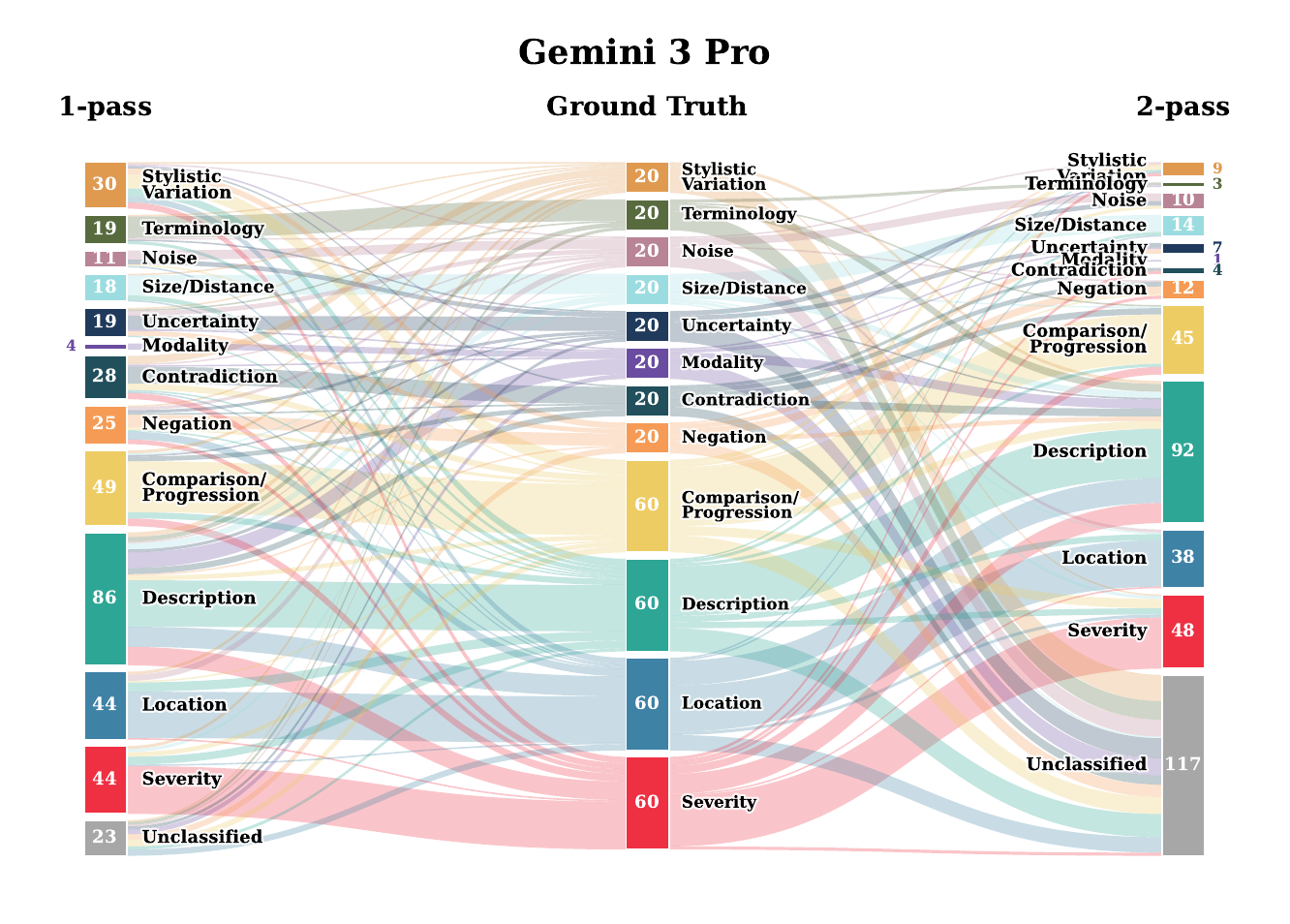} &
\includegraphics[width=0.47\textwidth]{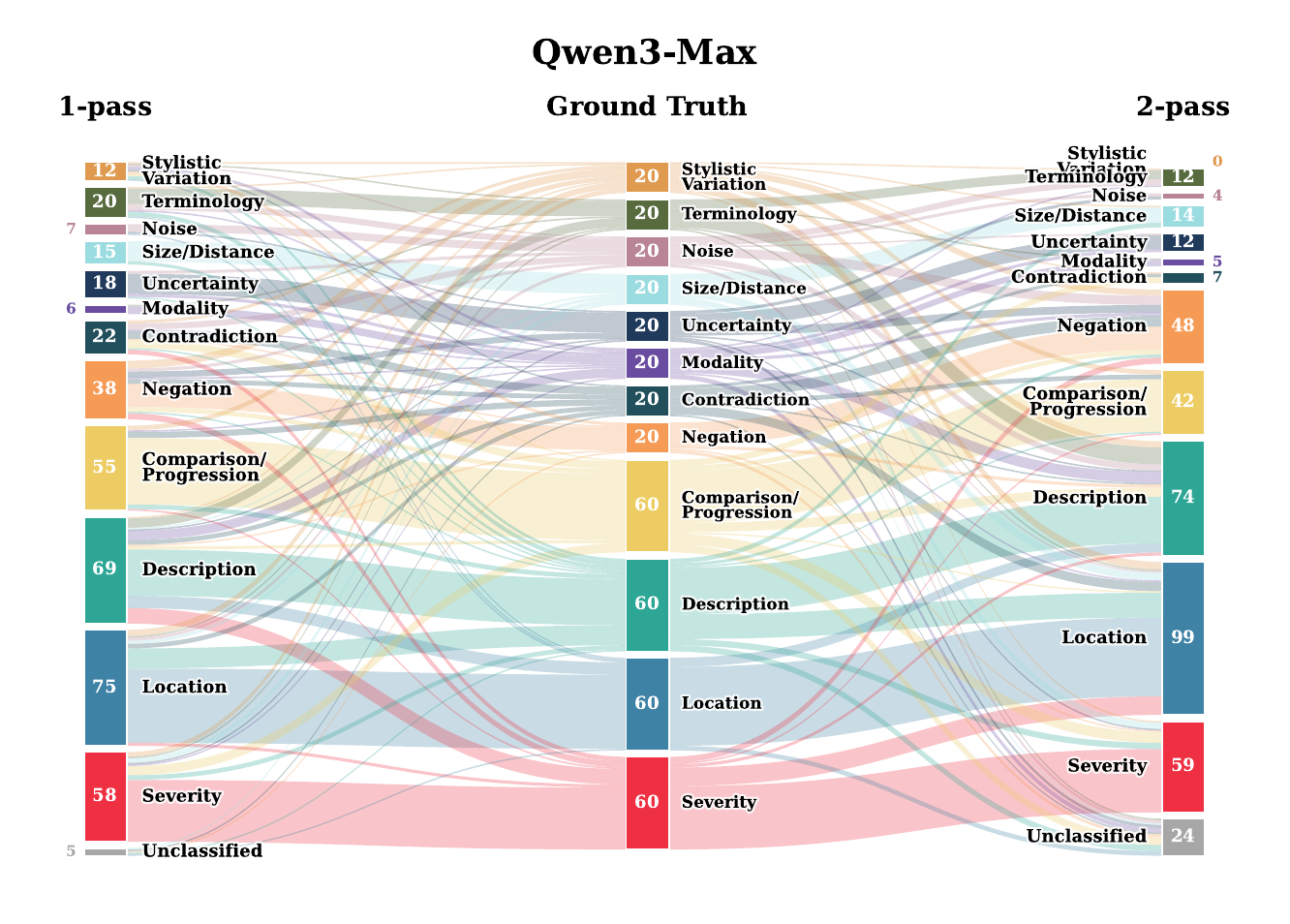} \\[-0.4em]

\includegraphics[width=0.47\textwidth]{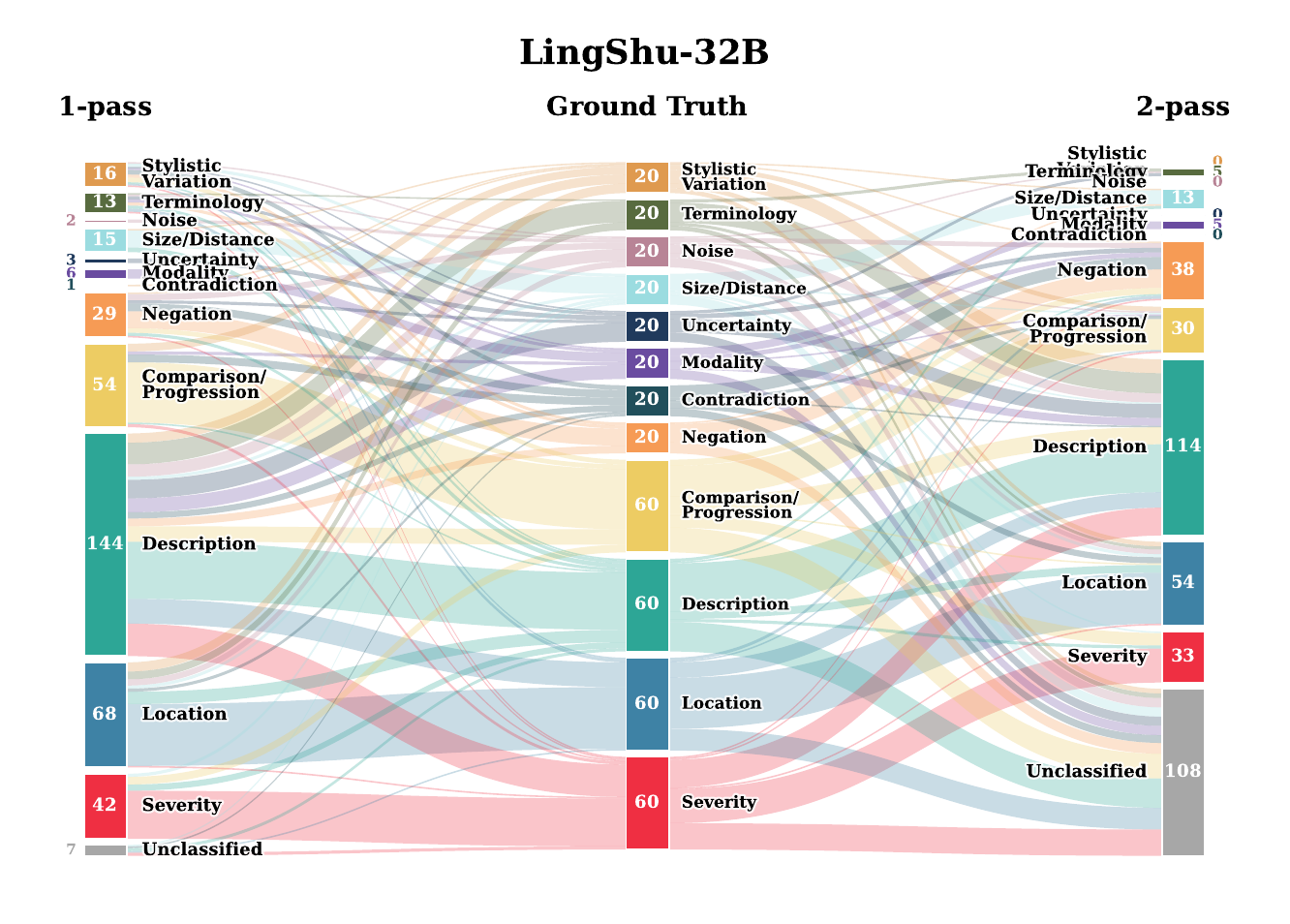} &
\includegraphics[width=0.47\textwidth]{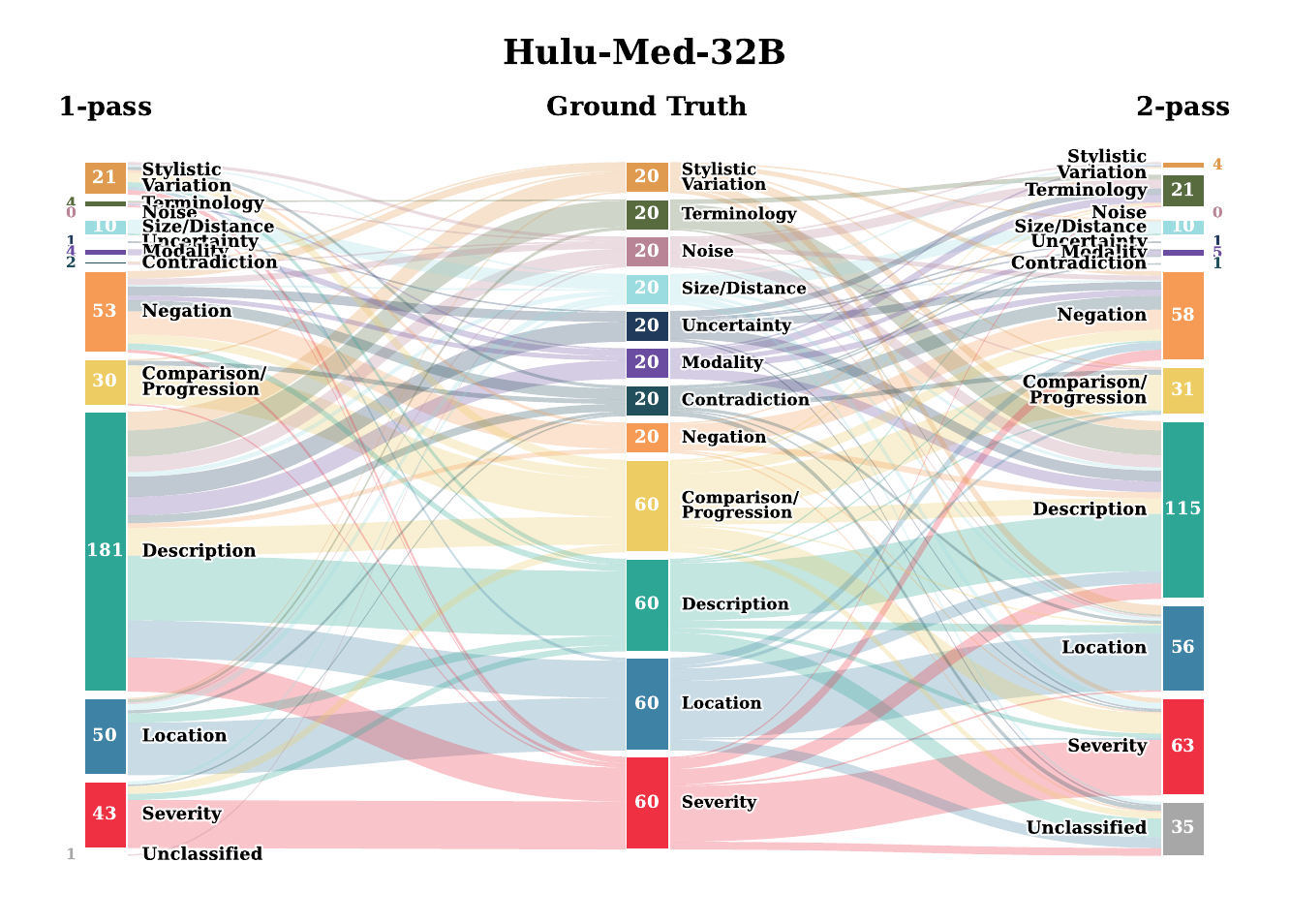} \\[-0.4em]

\includegraphics[width=0.47\textwidth]{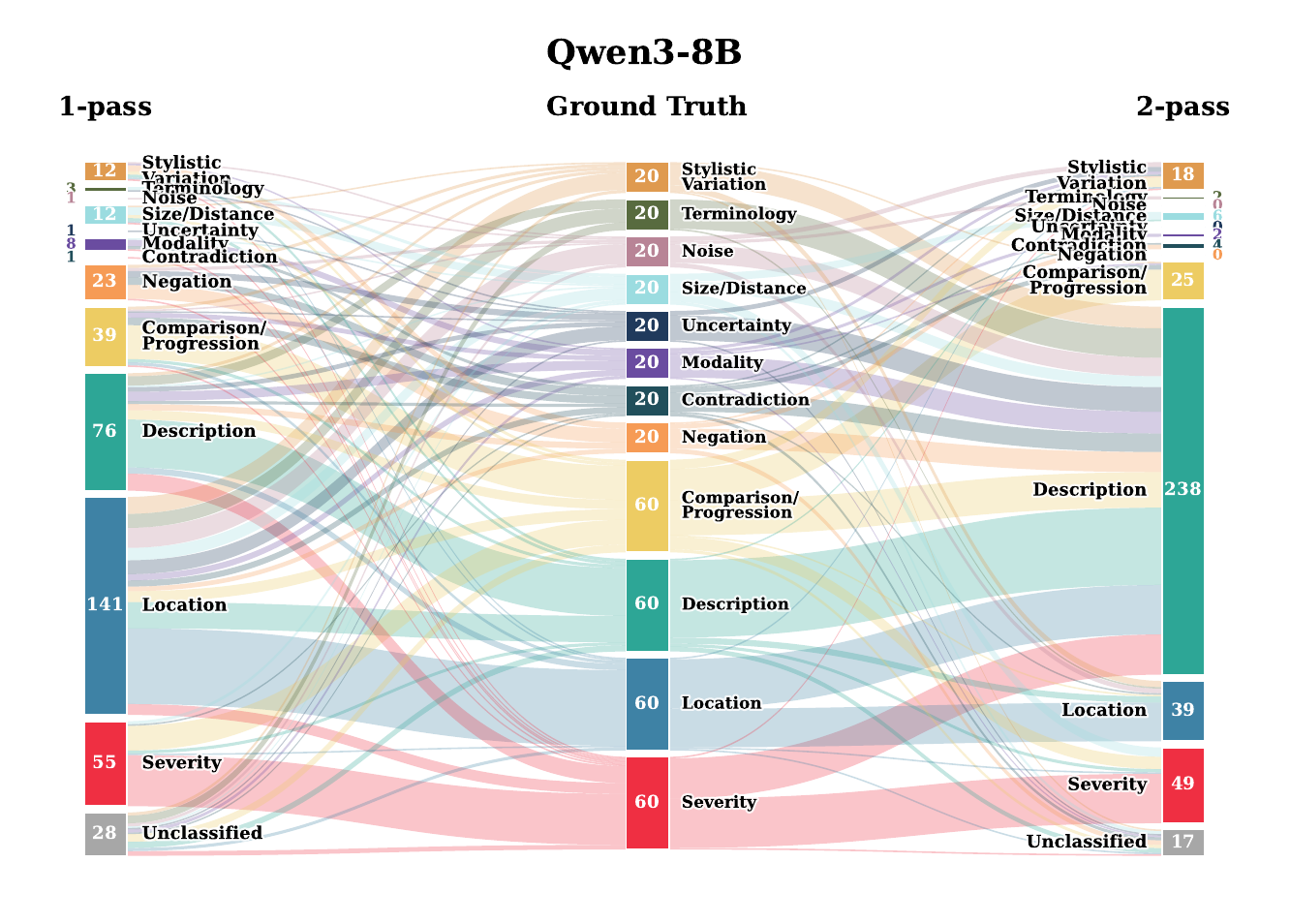} &
\includegraphics[width=0.47\textwidth]{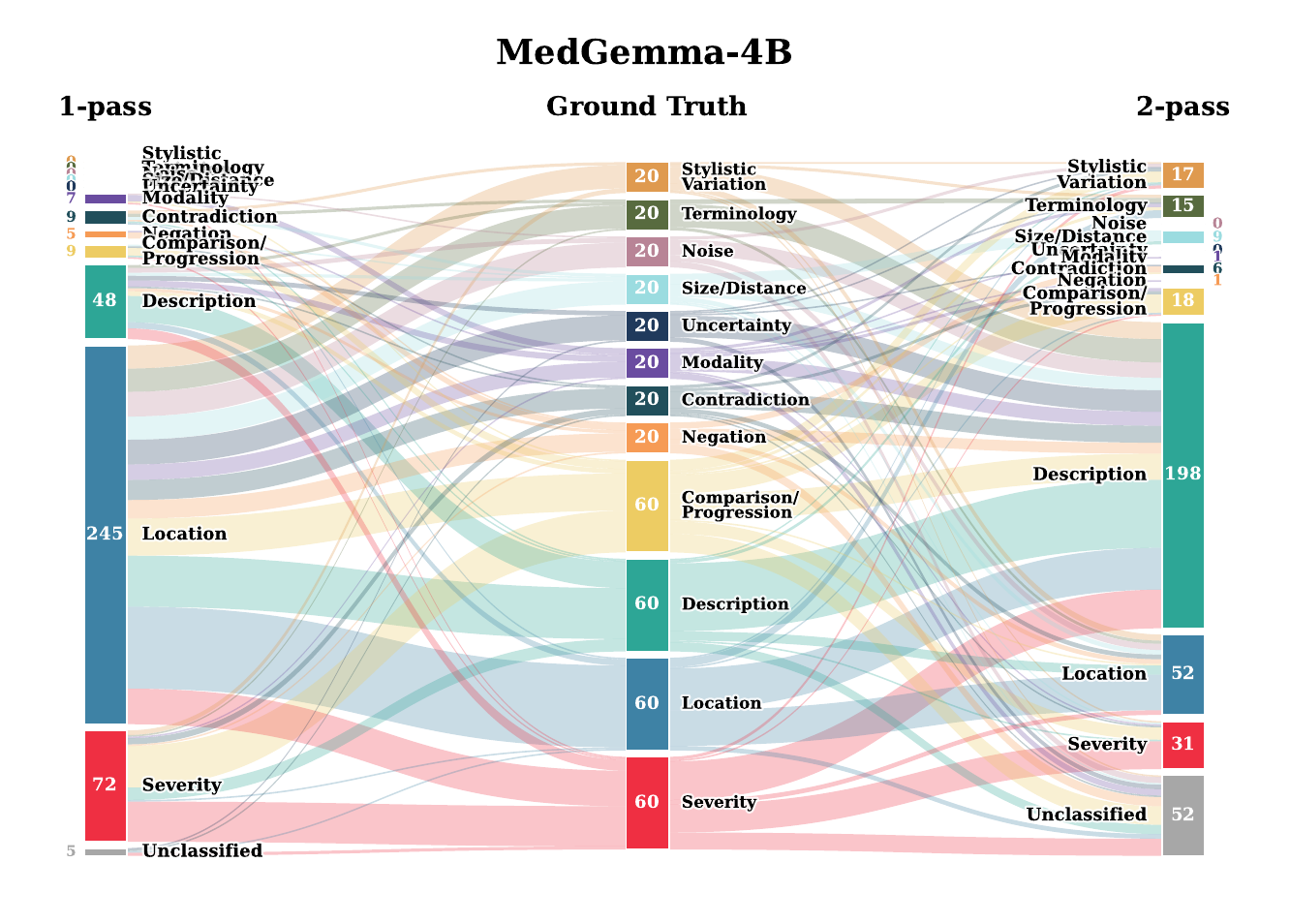}
\end{tabular}

\vspace{-0.5em}
\caption{
Aspect-flow Sankey diagrams across eight LLM evaluators. Each diagram shows
one-pass predicted aspects on the left, ground-truth aspects in the middle,
and two-pass predicted aspects on the right. Row totals are displayed on all
axes.
}
\label{fig:app_sankey_all}
\vspace{-0.8em}
\end{figure*}

\section{Discussion}
\label{app:discussion}

\subsection{Main Contribution and Practical Value}
\label{app:main_contribution_practical_value}

\paragraph{Contribution.}
Our contribution goes beyond applying ReEvalMed as a benchmark.
While ReEvalMed reports D and R mainly through averaged metric scores, we reformulate clinical-significance evaluation as a classification problem, which \textit{reduces the influence of outlier scores and makes the D--R behavior of LLM-as-judge metrics more explicit}.
This redefinition allows the discrimination bias of LLM evaluators to be measured more directly: a reliable evaluator should both identify clinically significant errors and tolerate insignificant variations.
A second contribution is showing that \textit{significance-oriented data augmentation can improve this balance}.
By synthesizing report pairs with controlled significance levels and applying a standard SFT--DPO post-training pipeline, we can calibrate lightweight LLM evaluators on the D--R boundary without designing a complex task-specific training algorithm.

\subsection{Same-Model Bias on Claude}
\label{app:same_model_bias}

Same-model bias is a potential concern in LLM-as-judge studies~\cite{zheng2023judging, liu-etal-2023-g}, especially when model-generated outputs are evaluated by the same or closely related model.
However, this concern does not apply to our main evaluation setting.
All main results, including Tables~\ref{tab:main} and~\ref{tab:training}, are evaluated on ReEvalMed, a clinically annotated benchmark that \textbf{is not generated by Claude.
Thus, our evaluation does not form a closed loop in which Claude-generated data are judged again by Claude.}
Claude is used only to construct supervision signals for data augmentation, which are then used to post-train lightweight Qwen and MedGemma models, not to optimize Claude itself.
Using a stronger model to generate supervision and transfer this capability to smaller models is a common distillation practice, and should not be interpreted as same-model bias in the evaluation.

To further reduce the possibility that model-generated errors are later judged by the same model, we used GPT-5.5 rather than Claude when constructing the Open-i evaluation set. GPT-5.5 is not included among the evaluator models in this paper. This design minimizes potential same-model bias in the external evaluation and further supports that our trained metrics do not merely imitate Claude's judgments, but improve their ability to distinguish clinically significant errors from insignificant variations.

\subsection{Potential Data Leakage}
\label{app:potential_data_leakage}

Potential data leakage from MIMIC-CXR is unlikely to directly affect the main conclusions of this work. Both the ReEvalMed test set and the samples constructed in our study are based on MIMIC-CXR reports with controlled injected clinical errors. Even if an LLM had been exposed to the original ground-truth MIMIC-CXR reports during pretraining, such exposure would not imply an ability to correctly identify diverse injected errors or determine their clinical significance, which is the central focus of our evaluation.

Moreover, ReEvalMed was released only at the end of 2025, whereas the evaluated LLMs were trained before that release, making direct benchmark leakage unlikely. To further reduce concerns about dataset-specific leakage, we additionally introduce an external Open-i evaluation set. Since Open-i is independent of MIMIC-CXR, this held-out evaluation provides complementary evidence that the observed behavior is not solely driven by memorization of MIMIC-CXR reports.

Although limited domain coverage remains a limitation, as discussed in the Limitation section, the added Open-i analysis helps assess whether the conclusions generalize beyond the primary MIMIC-CXR-derived setting.

\subsection{CRIMSON and MedGemma Fine-tuning}
\label{app:crimson_medgemma}

Compared with CRIMSON, our MedGemma-4B + SFT + RL evaluator changes the error profile from a more balanced score-based evaluator to a more robustness-oriented evaluator. CRIMSON achieves stronger discrimination accuracy (75.0 vs. 66.5), whereas our model achieves substantially higher robustness accuracy (90.5 vs. 80.5). As a result, the overall Avg is numerically similar, with our model slightly higher (78.5 vs. 77.8), but the paired significance test indicates that this Avg difference is not statistically significant. This suggests that the main gain of our approach is not a large increase in aggregate Avg over CRIMSON, but a shift toward better preservation of clinically insignificant variations.

The improvement is most visible on robustness cases, where our model correctly accepts more benign perturbations than CRIMSON. In particular, robustness accuracy improves substantially for fabrication-type cases (50.0 to 100.0), severity-related cases (73.3 to 96.7), contradiction cases (30.0 to 80.0), and terminology/style-related cases. This indicates that SFT + RL makes the evaluator less likely to over-penalize clinically harmless edits. However, this comes with a trade-off in discrimination: CRIMSON remains better at detecting significant errors overall, especially for size/distance, terminology, uncertainty, and severity-related significant errors. Thus, our method improves robustness and reduces false alarms on insignificant changes, while CRIMSON remains more sensitive to certain clinically significant discrepancies.

\subsection{Critical and Significant Severity Levels}
\label{app:critical_significant}

Our evaluation maps both \textit{critical} and \textit{significant} errors to the clinically significant class.
This design follows the practical goal of identifying report differences that may affect clinical interpretation or downstream decisions.
However, the boundary between critical and significant can itself be subjective and context-dependent.
For this reason, we treat the central task as distinguishing clinically meaningful from clinically insignificant differences, while leaving finer-grained severity calibration as an important direction for future work.

\subsection{Potential Error Propagation}
\label{app:error_propagation}

Our current two-pass design is itself motivated by common practice in machine translation evaluation. As pointed out in related work such as MQM-APE~\cite{lu-etal-2025-mqm}, in error span detection, completely missing an error span is relatively rare, while producing too many redundant spans is often a more common issue. Based on this observation, pass2 does not handle spans missed by pass1, but it can suppress redundant or unreasonable spans. We will clarify this design motivation and add the corresponding citation more explicitly in the revised version.

\subsection{Why Use a Standard Post-training Design}
\label{app:standard_post_training}

We use a standard SFT followed by DPO training design because our goal is to test whether targeted supervision can sharpen the clinical-significance boundary, rather than to introduce a new optimization algorithm.
SFT teaches the model the structured output format and exposes it to controlled clinical-error patterns.
DPO then directly optimizes preference pairs that reflect the desired D--R behavior.
This simple and reproducible design makes the effect of the data and supervision signal easier to interpret.

The same principle also applies to our prompting design.
Our prompts are intended to provide a clear and consistent evaluation protocol for testing whether the proposed clinical-significance formulation is effective, rather than to exhaustively optimize prompt engineering strategies.
A systematic exploration of alternative prompts, prompt ensembles, and model-specific prompting designs is therefore left for future work.

\section{Error Category Descriptions}
\label{app:error_categories}

Table~\ref{tab:error_desc} provides detailed descriptions of the error categories in ReEvalMed, organised by Error Aspect and Error Type.

\begin{table*}[h]
\centering
\footnotesize
\renewcommand{\arraystretch}{1.3}
\begin{tabular}{p{1.0cm}p{1.4cm}p{2.8cm}p{4.3cm}p{4.3cm}}
\toprule
\textbf{Type} & \textbf{Aspect} & \textbf{Description} & \textbf{Significant Example} & \textbf{Insignificant Example} \\
\midrule
O / F / E & Location & Errors in the precise anatomical site of a finding (e.g., laterality, lobe, region). &
  \textbf{REF}: \textit{left-sided rib fractures} \newline \textbf{TGT}: \textit{right rib fractures} &
  \textbf{REF}: \textit{left retrocardiac opacity} \newline \textbf{TGT}: \textit{opacity behind the heart on the left side} \\ \midrule
O / F / E & Severity & Errors in describing the extent or clinical seriousness of a finding. &
  \textbf{REF}: \textit{Heart is mildly enlarged} \newline \textbf{TGT}: \textit{Severely enlarged heart} &
  \textbf{REF}: \textit{Severe cardiomegaly} \newline \textbf{TGT}: \textit{Moderate-to-severe cardiomegaly} \\ \midrule
O / F / E & Description & Errors in morphological characteristics such as shape, margins, or appearance. &
  \textbf{REF}: \textit{irregular mass with spiculated margins} \newline \textbf{TGT}: \textit{round, smooth mass} &
  \textbf{REF}: \textit{patchy ill-defined opacities} \newline \textbf{TGT}: \textit{faint and poorly marginated opacities} \\ \midrule
O / F / E & Comp./Prog. & Errors in describing interval changes compared to prior imaging. &
  \textbf{REF}: \textit{Pulmonary edema has improved} \newline \textbf{TGT}: \textit{Pulmonary edema has worsened} &
  \textbf{REF}: \textit{No interval change in pleural effusion} \newline \textbf{TGT}: \textit{Pleural effusion is essentially unchanged} \\ \midrule
S & Negation & Incorrect presence or absence of a finding. &
  \textbf{REF}: \textit{No evidence of pneumothorax} \newline \textbf{TGT}: \textit{Pneumothorax is present} &
  \textbf{REF}: \textit{No pleural effusion is seen} \newline \textbf{TGT}: \textit{There is no definite pleural effusion} \\ \midrule
S & Modality & Conflicts with the imaging modality. &
  \textbf{REF}: \textit{Refer to prior CT torso for details} \newline \textbf{TGT}: \textit{Refer to prior abdominal ultrasound} &
  \textbf{REF}: \textit{Consider chest CT for further evaluation} \newline \textbf{TGT}: \textit{CT can be considered for further assessment} \\
S & Size/Distance & Errors in quantitative measurements. &
  \textbf{REF}: \textit{3-cm mass in the lingula} \newline \textbf{TGT}: \textit{8-cm mass in the lingula} &
  \textbf{REF}: \textit{ET tube within 1\,cm of the carina} \newline \textbf{TGT}: \textit{ET tube within 0.9\,cm of the carina} \\ \midrule
S & Contradiction & Internal logical inconsistencies within the same report. &
  \textbf{REF}: \textit{The lungs are clear.} \newline \textbf{TGT}: \textit{The lungs are clear. There is consolidation in the right base} &
  \textbf{REF}: \textit{No evidence of larger pleural effusions} \newline \textbf{TGT}: \textit{No evidence of larger pleural effusions. Minimal effusions may exist} \\ \midrule
S & Uncertainty & Incorrect use of hedging terms conveying diagnostic uncertainty. &
  \textbf{REF}: \textit{Whether this is pneumonia is radiographically indeterminate} \newline \textbf{TGT}: \textit{Pneumonia exists} &
  \textbf{REF}: \textit{A possible infiltrate is suggested} \newline \textbf{TGT}: \textit{An infiltrate is likely present} \\ \midrule
S & Terminology & Inaccurate or unclear medical terminology. &
  \textbf{REF}: \textit{A cavitary lesion, suggesting tuberculosis} \newline \textbf{TGT}: \textit{A hole, suggesting infection} &
  \textbf{REF}: \textit{A 3-cm mass} \newline \textbf{TGT}: \textit{A 3-cm lesion} \\ \midrule
S & Noise & Grammatical mistakes, typographical errors, or other linguistic noise. &
  \textbf{REF}: \textit{3-cm mass in the lingula has grown} \newline \textbf{TGT}: \textit{3-cm lingula margins has been growing irregularly} &
  \textbf{REF}: \textit{subtle opacity may represent atelectasis} \newline \textbf{TGT}: \textit{subtble opaciti may represent atelectasi} \\ \midrule
S & Stylistic Var. & Variations in phrasing that do not alter clinical meaning. &
  \textbf{REF}: \textit{Bilateral left greater than right pleural effusion} \newline \textbf{TGT}: \textit{Fluid accumulation on both sides of the chest, more on the right} &
  \textbf{REF}: \textit{lung\ldots{} pulmonary edema\ldots{} pleural effusions} \newline \textbf{TGT}: \textit{pleural effusions\ldots{} lung\ldots{} pulmonary edema} \\
\bottomrule
\end{tabular}
\renewcommand{\arraystretch}{1}
\caption{Descriptions and examples of the error categories in ReEvalMed~\cite{li2025reevalmed}, organized over 12 clinical aspects. Each aspect is associated with up to three error types: Omission~(O), Fabrication~(F), and Inaccuracy~(E). S: Inaccuracy, fabrication, and omission errors are randomly and evenly distributed.Each category appears in both the Discrimination and Robustness test sets. Examples are shown as REF--TGT pairs, where significant examples represent clinically meaningful errors and insignificant examples represent harmless variations.}
\label{tab:error_desc}
\end{table*}

\section{Length Distribution of Synthesized Data}
\label{app:length_dist}

\begin{figure*}[h]
\centering
\includegraphics[width=0.88\textwidth]{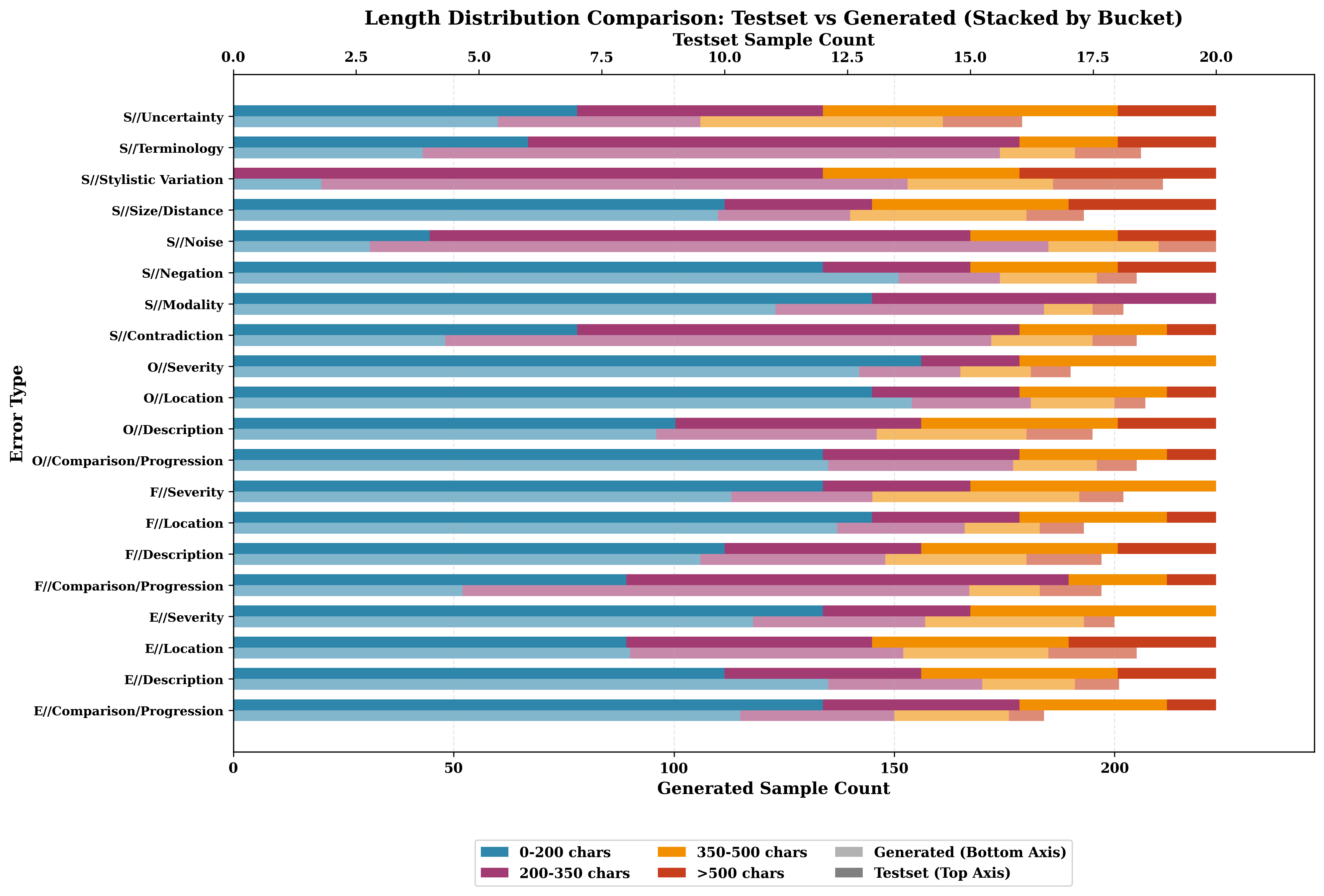}
\caption{Length distribution comparison between the ReEvalMed test set
(top axis) and our generated training corpus (bottom axis), stacked by
length bucket across all 12 aspects.}
\label{fig:length_dist}
\end{figure*}

Figure~\ref{fig:length_dist} shows the length distribution of our
synthesized data compared to the ReEvalMed test set, broken down by
error category and stratified across four length buckets.
The generated corpus covers all buckets in every category, confirming
that stratified sampling produces diverse report lengths.

\section{Human Validation on Report Synthesis}
\label{app:human_eval}

To verify the clinical fidelity of our synthesized data, we randomly sample 100 pairs (50~D + 50~R) covering the full range of error types and have a clinician manually review each pair.
For each sample, the clinician assesses whether:
(1)~the injected error is clinically plausible and correctly categorised,
(2)~the severity label (significant vs.\ insignificant) is appropriate,
and (3)~the generated report reads naturally without obvious artefacts.

\paragraph{Case studies}
Table~\ref{tab:human_eval_cases} presents representative examples
from the human validation, illustrating both successful and
problematic generations.

\begin{table*}[h]
\centering
\small
\renewcommand{\arraystretch}{1.2}
\begin{tabular}{p{15.5cm}}
\toprule
\textbf{Case 1} \enspace \colorbox{SevMislabel!15}{\textcolor{SevMislabel}{\scriptsize\textbf{Severity Mislabel}}} \hfill \textit{Significant / Noise} \\
\midrule
\textbf{REF:} ... There is no intraperitoneal free air. The lungs are clear without effusion or pneumothorax. ... IMPRESSION: Normal chest. \\
\textbf{TGT:} ... Severe motion artifact and noise significantly limit evaluation; intraperitoneal free air cannot be excluded. ... Recommend repeat imaging to exclude free air. \\
\hdashline\noalign{\vskip 2pt}
\textbf{Error Span:} ``no intraperitoneal free air'' $\to$ ``cannot be excluded'' \\
\textbf{Clinician:} Retake imaging only; no serious clinical consequence. Label should be \textit{insignificant}. \\
\midrule
\textbf{Case 2} \enspace \colorbox{SevMislabel!15}{\textcolor{SevMislabel}{\scriptsize\textbf{Severity Mislabel}}} \hfill \textit{Significant / Noise} \\
\midrule
\textbf{REF:} ... Frontal and lateral radiographs of the chest demonstrate normal heart size, mediastinal and hilar contours. Clear lungs. No pneumothorax or pleural effusion. IMPRESSION: Normal chest radiograph. \\
\textbf{TGT:} ... Severe motion artifact and excessive noise degrade image quality, limiting diagnostic evaluation. ... Pneumothorax or pleural effusion cannot be excluded. IMPRESSION: Nondiagnostic chest radiograph ... Recommend repeat imaging ... \\
\hdashline\noalign{\vskip 2pt}
\textbf{Error Span:} ``normal heart size ... Clear lungs'' $\to$ ``Severe motion artifact ... cannot be reliably assessed'' \\
\textbf{Clinician:} Same pattern as Case~1; normal study changed to nondiagnostic with repeat imaging recommendation. No actual pathology introduced. Label should be \textit{insignificant}. \\
\midrule
\textbf{Case 3} \enspace \colorbox{InjectFail!15}{\textcolor{InjectFail}{\scriptsize\textbf{Injection Failure}}} \hfill \textit{Significant / Location-Omission} \\
\midrule
\textbf{REF:} CHEST ON \_\_ HISTORY: Pneumonitis. REFERENCE EXAM: \_\_. Compared to the prior exam, there is no significant interval change. \\
\textbf{TGT:} CHEST ON \_\_ HISTORY: Pneumonitis. REFERENCE EXAM: \_\_. Compared to the prior exam, there is no significant interval change in the pneumonitis. \\
\hdashline\noalign{\vskip 2pt}
\textbf{Error Span:} \textit{null} $\to$ ``in the pneumonitis'' \\
\textbf{Clinician:} Appears to critique the REF rather than inject an error into the TGT; the added location specificity is arguably an improvement rather than an omission. \\
\midrule
\textbf{Case 4} \enspace \colorbox{InjectFail!15}{\textcolor{InjectFail}{\scriptsize\textbf{Injection Failure}}} \hfill \textit{Significant / Location-Omission} \\
\midrule
\textbf{REF:} HISTORY: Hickman catheter placement. FINDINGS: The catheter tip lies at the level of the mid portion of the SVC. No evidence of acute pneumonia or vascular congestion. \\
\textbf{TGT:} \textit{(identical to REF)} \\
\hdashline\noalign{\vskip 2pt}
\textbf{Error Span:} \textit{null} $\to$ \textit{null} \\
\textbf{Clinician:} Error injection failed entirely; REF and TGT are identical. The model's explanation reveals an aborted attempt: ``\textit{This is actually appropriate positioning. Let me generate a proper SEVERE omission error.}'' \\
\midrule
\textbf{Case 5} \enspace \colorbox{AspectMis!15}{\textcolor{AspectMis}{\scriptsize\textbf{Aspect Mismatch}}} \hfill \textit{Insignificant / Negation} \\
\midrule
\textbf{REF:} ... IMPRESSION: In comparison with the study of \_\_ \\
\textbf{TGT:} ... IMPRESSION: Compared with the study of \_\_ \\
\hdashline\noalign{\vskip 2pt}
\textbf{Error Span:} ``In comparison with'' $\to$ ``Compared with'' \\
\textbf{Clinician:} Purely stylistic; the model failed to inject the assigned \textit{Negation} error. \\
\midrule
\textbf{Case 6} \enspace \colorbox{AspectMis!15}{\textcolor{AspectMis}{\scriptsize\textbf{Aspect Mismatch}}} \hfill \textit{Insignificant / Contradiction} \\
\midrule
\textbf{REF:} ... There is no large pleural effusion or pneumothorax. ... \\
\textbf{TGT:} ... No large pleural effusion or pneumothorax is identified. ... \\
\hdashline\noalign{\vskip 2pt}
\textbf{Error Span:} ``There is no large pleural effusion or pneumothorax.'' $\to$ ``No large pleural effusion or pneumothorax is identified.'' \\
\textbf{Clinician:} Purely stylistic; the model failed to inject the assigned \textit{Contradiction} error. \\
\midrule
\textbf{Case 7} \enspace \colorbox{AspectMis!15}{\textcolor{AspectMis}{\scriptsize\textbf{Aspect Mismatch}}} \hfill \textit{Insignificant / Size-Distance} \\
\midrule
\textbf{REF:} ... IMPRESSION: AP and lateral chest compared to \_\_: Normal heart, lungs, hila, mediastinum, and pleural surfaces. \\
\textbf{TGT:} ... IMPRESSION: AP and lateral chest compared to \_\_: Normal heart size, lungs, hila, mediastinum, and pleural surfaces. \\
\hdashline\noalign{\vskip 2pt}
\textbf{Error Span:} ``Normal heart'' $\to$ ``Normal heart size'' \\
\textbf{Clinician:} Purely stylistic; the model failed to inject the assigned \textit{Size/Distance} error. \\
\bottomrule
\end{tabular}
\renewcommand{\arraystretch}{1}
\caption{Representative cases from human validation of synthesized data. Cases are categorised by failure type: \colorbox{SevMislabel!15}{\textcolor{SevMislabel}{\scriptsize\textbf{Severity Mislabel}}} (severity label disagrees with clinical judgement), \colorbox{InjectFail!15}{\textcolor{InjectFail}{\scriptsize\textbf{Injection Failure}}} (error injection produces no meaningful change), and \colorbox{AspectMis!15}{\textcolor{AspectMis}{\scriptsize\textbf{Aspect Mismatch}}} (injected change is purely stylistic, deviating from the assigned error type).}
\label{tab:human_eval_cases}
\end{table*}

\section{Score-based Metric Results}
\label{app:curves}

Unlike LLM-based metrics that output binary significant/insignificant
predictions, score-based metrics produce continuous scores requiring a
decision threshold to convert to D/R accuracy.
In the main paper (Table~\ref{tab:main}), we report results at the
\textbf{maximin threshold} $\max_\theta \min(D_\theta, R_\theta)$,
which selects the operating point that maximises the worse of the two
dimensions.

\begin{figure*}[t]
\centering
\includegraphics[width=1.8\columnwidth]{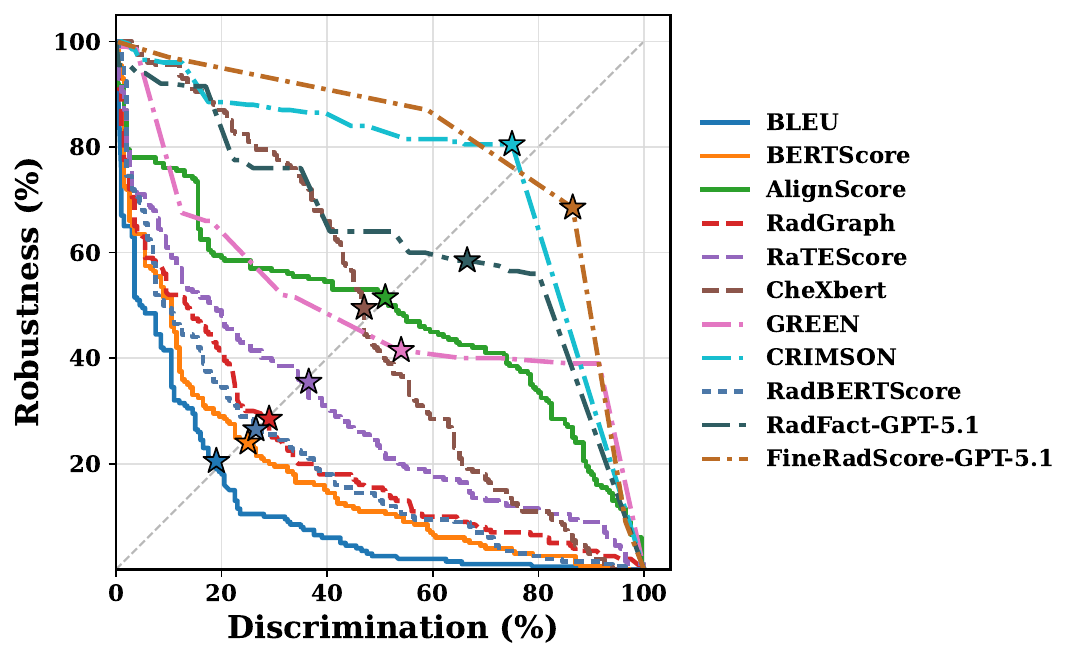}
\caption{D--R trade-off curves for all score-based metrics.
Each curve traces (D, R) accuracy across decision thresholds;
filled circles mark the maximin operating point
$\max_\theta \min(D_\theta, R_\theta)$ reported in Table~\ref{tab:main}.}
\label{fig:dr_curve_appendix}
\end{figure*}

Figure~\ref{fig:dr_curve_appendix} shows the full D--R trade-off
curves for all 8 score-based metrics.
Each curve traces the (D, R) accuracy pair as the decision threshold
varies; the filled circle marks the maximin operating point reported in
Table~\ref{tab:main}.
CRIMSON achieves the best maximin point (75.0, 80.5), substantially
outperforming all other score-based metrics.
NLP metrics (BLEU, BERTScore, AlignScore) cluster in the low-accuracy
region, while medical metrics (RadGraph, RaTEScore, CheXbert) offer
moderate improvements.

\section{Error Specification Examples}
\label{app:error_spec}

We provide two representative error specifications generated by
Claude Sonnet 4.5 (Section~\ref{sec:data}).
Each specification contains a clinical definition, medical patterns,
and contrastive text patterns that distinguish significant errors
from harmless variations.

\begin{tcolorbox}[
colback=white,
colframe=black!60,
title=Error Specification: Location -- Omission,
breakable]

\textbf{Definition:}
Omission of anatomical location information present in the reference report, including missing specific anatomical sites, laterality (left/right), spatial relationships, or regional descriptors.

\medskip
\textbf{Medical Patterns:}
\begin{itemize}[leftmargin=*,topsep=2pt,itemsep=1pt]
  \item Missing laterality specification (left/right, bilateral)
  \item Missing specific anatomical region or lobe identification
  \item Missing spatial descriptors (upper/lower, anterior/posterior)
\end{itemize}

\medskip
\textbf{Discrimination Text Patterns} (significant):
\begin{itemize}[leftmargin=*,topsep=2pt,itemsep=1pt]
  \item ``left upper lobe nodule'' $\to$ ``pulmonary nodule'' (missing critical laterality and lobe)
  \item ``bilateral pleural effusions'' $\to$ ``pleural effusion'' (missing bilateral nature)
  \item ``fracture of L3 vertebral body'' $\to$ ``lumbar fracture'' (missing specific vertebral level)
\end{itemize}

\medskip
\textbf{Robustness Text Patterns} (insignificant):
\begin{itemize}[leftmargin=*,topsep=2pt,itemsep=1pt]
  \item ``hepatic lesion in segment 7'' $\to$ ``right hepatic lobe lesion'' (alternative valid descriptor)
  \item ``distal esophagus'' $\to$ ``lower esophagus'' (synonymous location terms)
  \item ``left lung base'' $\to$ ``left lower lung'' (equivalent regional description)
\end{itemize}
\end{tcolorbox}

\begin{tcolorbox}[
colback=white,
colframe=black!60,
title=Error Specification: Description -- Fabrication,
breakable]

\textbf{Definition:}
Adding findings, observations, or anatomical descriptions in TGT that do not exist in REF, including introducing new pathological findings, normal structures not mentioned, or additional descriptive details.

\medskip
\textbf{Medical Patterns:}
\begin{itemize}[leftmargin=*,topsep=2pt,itemsep=1pt]
  \item Adding new pathological findings not present in REF (e.g.\ pleural effusion, nodules)
  \item Adding quantitative details or measurements that do not exist in REF
  \item Introducing temporal or comparative information not in REF
\end{itemize}

\medskip
\textbf{Discrimination Text Patterns} (significant):
\begin{itemize}[leftmargin=*,topsep=2pt,itemsep=1pt]
  \item REF: ``clear lungs'' $\to$ TGT adds ``bilateral pleural effusions''
  \item REF: ``normal heart'' $\to$ TGT adds ``cardiomegaly''
  \item REF: ``opacity present'' $\to$ TGT adds ``large 5cm mass''
\end{itemize}

\medskip
\textbf{Robustness Text Patterns} (insignificant):
\begin{itemize}[leftmargin=*,topsep=2pt,itemsep=1pt]
  \item REF: ``no acute findings'' $\to$ TGT: ``heart, lungs, and mediastinum show no acute findings''
  \item REF: ``unremarkable'' $\to$ TGT lists ``bones intact, soft tissues normal''
  \item REF gives findings $\to$ TGT adds ``adequate inspiration'' as contextual information
\end{itemize}
\end{tcolorbox}

\section{Prompt Templates}
\label{app:prompt}

We provide the full prompt templates used for inference
(Section~\ref{sec:benchmark}) and data synthesis
(Section~\ref{sec:data}).
In all templates, \hl{\{REF\}} and \hl{\{TGT\}} are replaced
with the ground-truth and target reports, respectively.


\hypertarget{box:onepass}{}
\begin{tcolorbox}[
colback=white,
colframe=OnePassColor,
title=One-Pass Evaluation Prompt,
breakable]

\textbf{SYSTEM}:

You are a clinical evaluation agent for chest X-ray reports.
You will be given:
1) a ground-truth report written by professional radiologists, and
2) a target report to be evaluated.

Your task is to:
\begin{itemize}[leftmargin=*,topsep=2pt,itemsep=1pt]
  \item Identify all clinically significant and insignificant errors in the target report.
  \item Classify each error into one of the predefined error categories.
  \item Determine the error type (Omission / Fabrication / Inaccuracy) when applicable.
  \item Provide a concise explanation of the errors and their classifications.
\end{itemize}

Follow all definitions and constraints provided in subsequent instructions.

\medskip
\textbf{USER}:

You must evaluate the target report against the ground-truth report and produce a structured JSON output.

Your output must be a JSON object with the following structure:
\begin{itemize}[leftmargin=*,topsep=2pt,itemsep=1pt]
  \item ``critical'': a dictionary mapping error spans to error aspects for critical errors
  \item ``significant'': a dictionary mapping error spans to error aspects for significant errors
  \item ``insignificant'': a dictionary mapping error spans to error aspects for insignificant errors
  \item ``explanation'': a concise explanation of the errors identified and the rationale for the categorization
\end{itemize}

\textbf{Definition of significance:}
\begin{itemize}[leftmargin=*,topsep=2pt,itemsep=1pt]
  \item \textbf{Critical}: Refer to internal inconsistencies or logical contradictions, such as describing the presence of a structure previously stated to be absent, that can severely undermine clinician trust in the report and are considered critical failures.
  \item \textbf{Significant}: Errors that meaningfully alter, mislead, or distort clinical decision-making.
  \item \textbf{Insignificant}: Errors that represent stylistic variation, minor wording changes, or clinically harmless deviations.
\end{itemize}

\textbf{Error Aspects:}

Below are the 12 error categories and their descriptions:

\hl{\{error\_aspect\_info\}}

Every error must be assigned exactly one of these aspects.

\textbf{Error Types:}

For the ``Comparison/Progression'', ``Description'', ``Location'', ``Severity'' aspects, you must further specify the error type using one of:

\hl{\{error\_type\_info\}}

For all other error aspects, you must NOT include an error type; only the aspect label should be provided.

Example:
\begin{itemize}[leftmargin=*,topsep=2pt,itemsep=1pt]
  \item ``rib fractures'': ``Location - Omission''
  \item ``no pneumothorax'': ``Negation'' (no error type needed)
\end{itemize}

If no errors are identified, the ``significant'' and ``insignificant'' fields should be empty dictionaries.

\medskip
Reports to evaluate:

Ground-Truth Report: \hl{\{REF\}}\\
Target Report: \hl{\{TGT\}}

\medskip
Please output only a single valid JSON object.

IMPORTANT:
\begin{itemize}[leftmargin=*,topsep=2pt,itemsep=1pt]
  \item Do NOT wrap the JSON in Markdown code fences.
  \item Do NOT include any language labels such as ``json'' before the JSON.
  \item The output must start directly with \{\{ and end with \}\}.
  \item Do NOT include any extra text, comments, or explanations outside the JSON.
\end{itemize}

\end{tcolorbox}


\hypertarget{box:twopass_p1}{}
\begin{tcolorbox}[
colback=white,
colframe=TwoPassColor,
title=Two-Pass: Pass~1 (Span Detection),
breakable]

\textbf{SYSTEM}:

You are a clinical evaluator for chest X-ray reports.
Given a ground-truth report (REF) and a target report (TGT), identify all differences.

For each difference, output a JSON object with:
\begin{itemize}[leftmargin=*,topsep=2pt,itemsep=1pt]
  \item ``span'': the exact text span that changed (from TGT for Fabrication/Inaccuracy/Style; from REF for Omission)
  \item ``aspect'': one of the 12 error aspects
\end{itemize}

Valid aspect codes:
Location - Omission, Location - Fabrication, Location - Inaccuracy,
Severity - Omission, Severity - Fabrication, Severity - Inaccuracy,
Description - Omission, Description - Fabrication, Description - Inaccuracy,
Comparison/Progression - Omission, Comparison/Progression - Fabrication, Comparison/Progression - Inaccuracy,
Negation, Modality, Size/Distance, Contradiction, Uncertainty, Terminology, Noise, Stylistic Variation

Output a JSON array. If no differences, output [].
Do NOT wrap in markdown. Start directly with [ and end with ].

\medskip
\textbf{USER}:

Ground-Truth Report:\\
\hl{\{REF\}}

Target Report:\\
\hl{\{TGT\}}

Identify all differences and output the JSON array of \{span, aspect\} objects.

\end{tcolorbox}


\hypertarget{box:twopass_p2}{}
\begin{tcolorbox}[
colback=white,
colframe=TwoPassColor,
title=Two-Pass: Pass~2 (Severity Judgment),
breakable]

\textbf{SYSTEM}:

You are a clinical evaluator for chest X-ray reports.
You will be given a REF/TGT report pair, a specific error span, its aspect, and judgment criteria.
Determine whether the error is SIGNIFICANT or INSIGNIFICANT.

Output a single JSON object:
\{``severity'': ``significant''/``insignificant''/``critical'', ``explanation'': ``...''\}

Do NOT wrap in markdown. Start with \{ and end with \}.

\medskip
\textbf{USER}:

Ground-Truth Report:\\
\hl{\{REF\}}

Target Report:\\
\hl{\{TGT\}}

\rule{\linewidth}{0.4pt}
Identified Error Span: ``\hl{\{span\}}''\\
Error Aspect: \hl{\{aspect\}}\\
Error Type: \hl{\{etype\_label\}}

\rule{\linewidth}{0.4pt}
Judgment Criteria for [\hl{\{aspect\}}]:\\
\hl{\{criteria\}}

\rule{\linewidth}{0.4pt}
Is this error SIGNIFICANT or INSIGNIFICANT? Output only the JSON.

\end{tcolorbox}


\begin{tcolorbox}[
colback=white,
colframe=black!60,
title=Error Injection Prompt (Data Synthesis),
breakable]

\textbf{SYSTEM}:

You are a medical imaging report error generation system specialized in radiology reports.

\medskip
\textbf{USER}:

\textbf{Error Type Definition:}

Error Category: \hl{\{error\_type\}} (\hl{\{error\_category\_name\}})\\
Error Aspect: \hl{\{error\_aspect\}}

\hl{\{error\_definition\}}

\medskip
\textbf{Reference Medical Context:}

The following patterns show how this error type manifests in real medical reports:

\hl{\{medical\_patterns\}}

\medskip
\textbf{Severity Level:} \hl{\{severity\_description\}}

\textbf{Example Text Patterns:}

\hl{\{selected\_text\_patterns\}}

\medskip
\textbf{Task:}

\hl{\{severity\_instruction\}}

Reference Report: \hl{\{REF\}}

\textbf{Instructions:}
\begin{itemize}[leftmargin=*,topsep=2pt,itemsep=1pt]
  \item Introduce the specified error type into the reference report
  \item Follow the text patterns shown above as guidance
  \item Keep the rest of the report unchanged
  \item Maintain medical plausibility and report structure
  \item Preserve the overall report length ($\pm$20\%)
  \item Ensure the severity matches the specified level
\end{itemize}

Output ONLY the modified target report text, without any explanations or formatting.

\end{tcolorbox}

\end{document}